\documentclass[]{moworld}
\microtypesetup{expansion=false}
\usepackage{amsmath,amssymb,tabularx,array}
\usepackage{url}
\usepackage{fontawesome5}
\usepackage{xcolor}

\hypersetup{pdftitle={BWM-Forge: Agentic World Programming for Embodied Intelligence},pdfauthor={}}
\title{BWM-Forge: Agentic World Programming for\\ Embodied Intelligence}
\author{%
  BWM Team\\
  Tongji University \& Koala Uran\\[4pt]
  \faGlobe\ \url{https://bwm-forge.github.io/}
  \quad
  \faGithub\ \url{https://github.com/Lcixs/BWM-Forge}%
}
\abstract{%
Generating diverse robot videos requires control over scene geometry, robot motion, and viewpoints. End-to-end video generation offers limited direct access to these variables, making targeted edits and reuse across tasks difficult. We present \mbox{BWM-Forge}, a code-driven framework for controllable video generation in robotics. Given robot assets and a task description, the framework accepts an initial image, an image with motion data, or a human demonstration video. A coding agent constructs editable Blender scene programs that specify objects, robot motion, and cameras. The resulting proxy videos provide structural and motion references for a video generation model, which adds visual detail. A multi-agent harness coordinates input processing, scene construction, review, and rendering. Automated evaluation and user feedback support iterative refinement, while an experience repository stores and retrieves reusable modeling and rendering rules. This design supports multiple robot embodiments, synchronized viewpoints, and appearance variation from a shared scene program. We evaluate the framework on 15 tasks covering five robot types. Compared with \mbox{LingBot-World Base Cam} on these tasks, \mbox{BWM-Forge} achieves higher aesthetic quality, imaging quality, subject and background consistency, motion smoothness, and RoboJudge-9B physical adherence and instruction alignment scores. The baseline scores higher in dynamic degree. Qualitative cases illustrate viewpoint control, scene variation, feedback-driven edits, and experience reuse across scenes. These results demonstrate a practical approach to generating controllable robot videos through editable, executable scene representations.
}
\begin{document}
\maketitle
\vspace*{-12mm}
\vfill
\noindent\begin{minipage}{\textwidth}
  \centering
  \setlength{\parskip}{0pt}
  \includegraphics[width=0.86\linewidth]{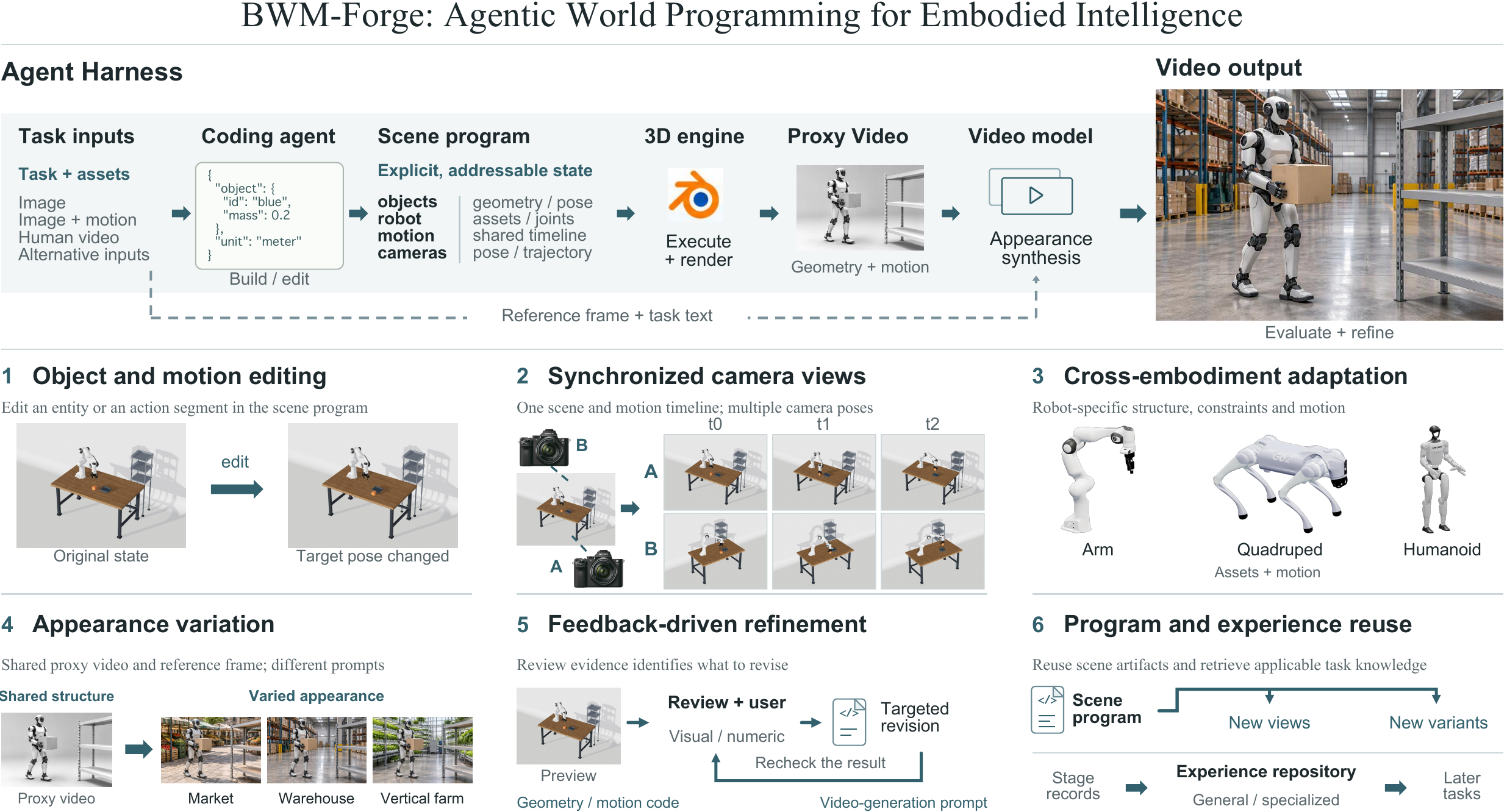}
  \captionsetup{format=plain,justification=centering,skip=5pt}
  \captionof*{figure}{\mbox{BWM-Forge} generates controllable robot videos using editable scene programs and proxy videos.}
\end{minipage}\par\vspace*{5pt}
\clearpage
\section{Introduction}
Video provides a natural representation of physical-world dynamics, motivating video world models that predict future visual observations. When world state is represented implicitly in generated frames, however, users have limited direct access to scene objects, robot configurations, and camera trajectories for editing and reuse. For embodied data generation, these limitations concern changing a task object while preserving the surrounding scene, observing the same interaction from another camera, and adapting an existing scene to a different robot. Recent code world models offer a complementary approach. WorldCoder and subsequent work represent world dynamics as executable programs for prediction and decision making \citep{tang2024worldcoder,lehrach2026codeworldmodels}. Code World Model and Code2Worlds extend this direction toward code-mediated visual generation and the construction of 3D scenes and motion \citep{chen2026codeworldmodel,zhang2026code2worlds}. These approaches motivate an explicit, executable representation through which scene structure and motion can be inspected, modified, and rendered.

\begin{figure}[!t]
\centering
\includegraphics[width=\linewidth]{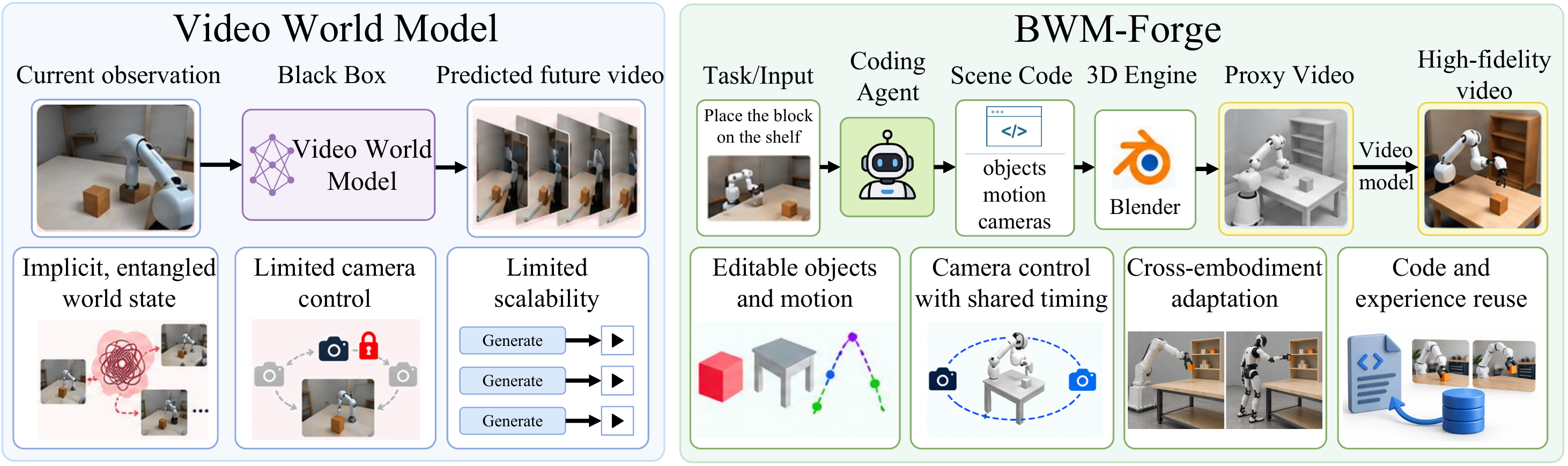}
\caption{Code-driven embodied video generation. Left: direct video prediction. Right: \mbox{BWM-Forge} uses editable scene programs and proxy videos to guide appearance generation, supporting scene edits, synchronized viewpoints, and reuse across robot embodiments.}
\label{fig:motivation}
\end{figure}

Inspired by this direction, we introduce \textbf{\mbox{BWM-Forge}}, a framework for \textit{agentic world programming} that generates controllable videos for embodied intelligence (Figure\nobreakspace{}\ref{fig:motivation}). In \mbox{BWM-Forge}, world programming means expressing scene objects, spatial relationships, robot motion, and camera states as executable scene programs; agents construct, execute, inspect, and revise these programs using tools and feedback. The goal is to generate diverse visual data across tasks, robot embodiments, and viewpoints, motivated by the need for broad robot interaction coverage reflected in datasets such as DROID and BridgeData V2 \citep{khazatsky2024droid,walke2023bridgedata}. A unified pipeline accepts robot assets and task descriptions together with an initial image, an image with motion data, or a human demonstration video, and produces high-fidelity videos of interactions between robots and their environments. The same workflow covers robotic arms, humanoid robots, quadrupeds, and dexterous hands by loading their corresponding assets and using the available motion evidence. These videos provide potential training material for embodied video generation and world modeling.

Prior work uses language models to generate robotic simulation tasks and scenes \citep{wang2024gensim,wang2024robogen}, while robot-conditioned video generation uses explicit robot structure or rendered conditions to guide visual synthesis \citep{kim2026robotfactored,huang2026roborender}. \mbox{BWM-Forge} combines executable scene construction with appearance generation through proxy videos as an intermediate representation. A coding agent controls Blender to construct task scenes, import robot description packages for different embodiments, and coordinate robot motion and camera trajectories. Scene geometry, motion, and viewpoints remain directly editable in the program. Loading a different robot description package adapts the model to the target embodiment's structure, while the scene program retains editable objects and camera parameters. A 3D engine executes the program and renders proxy videos, which provide structural and motion guidance for a video generation model to add visual detail.

This representation supports interactive refinement and flexible viewpoint control. The coding agent can modify scene code and parameters in response to user feedback, including object placement, robot motion, and camera settings. Multiple fixed cameras produce synchronized views of the same task, while camera trajectories enable continuously varying viewpoints. Because proxy videos share a scene program and motion timeline, their geometric relationships and motion remain temporally aligned. These shared references guide subsequent high-fidelity generation. This provides complementary observations of the same interaction for learning robot motion and its relationship to the environment. Reusing the scene program with different robot assets or camera configurations provides a practical basis for expanding visual observations across embodiments and viewpoints.

The agentic workflow is implemented by a Harness that coordinates four specialized agents. The Preprocessing Agent parses requests, validates robot assets and motion data, and extracts visual evidence. The Building Agent constructs and animates scenes through code. The Review Agent inspects visual and numerical evidence, identifies defects, and applies local corrections. The Rendering Agent uses the resulting proxy videos, reference images, and prompts to generate and evaluate high-fidelity outputs. Automated checks and optional user feedback guide revisions to both proxy and rendered videos. Stage summaries and execution traces supply candidate experience entries, which the Harness consolidates and retrieves for later tasks. This cycle of programming, execution, review, and experience reuse connects the framework's editable world representation to its embodied data generation goal.

\begin{samepage}
Our main contributions are as follows:

\begin{itemize}

\item \textbf{Agentic world programming for embodied data generation.} We introduce \mbox{BWM-Forge}, a framework that converts multimodal task inputs into executable scene programs and uses their rendered outputs to guide high-fidelity robot video generation.

\item \textbf{Explicit control over scenes, embodiments, and viewpoints.} Editable scene programs expose object geometry, robot motion, and camera states for targeted revision, synchronized multi-view rendering, and adaptation to different robot assets.

\item \textbf{A collaborative generation and refinement workflow.} A four-agent Harness combines input validation, scene construction, evidence-based review, rendering, user feedback, and reusable experience to support iterative video generation across tasks.

\end{itemize}
\end{samepage}

\section{Related Work}
\subsection{Code World Model}

Current Code World Model approaches use Coding Agents to build executable worlds, compute state changes through code, and generate visual results with rendering models to better reflect physical laws. Early uses of coding focused on games and task planning, including state transitions, game rules, action validity checks, and simulations of future outcomes for search and decision making. For example, WorldCoder \citep{tang2024worldcoder} builds world models with Python programs for state prediction and decision making in grid worlds and planning tasks; Code World Models for General Game Playing \citep{lehrach2026codeworldmodels} converts game rules into executable code and uses Monte Carlo tree search for decision making. Code World Model: Coding Agent as World Brain \citep{chen2026codeworldmodel} further maintains world states through code, converts them into proxy videos, and uses a video generation model for realistic rendering. Later work extends coding to the construction and maintenance of 3D scenes, objects, cameras, and motion, as in Code2Worlds \citep{zhang2026code2worlds}, which generates Blender scenes and animation scripts with a Coding Agent and refines them through visual feedback. Coding also supports physical inference by proposing mechanisms from video, building executable models, and revising them based on differences between simulation and observation; Code as Worlds \citep{wang2026codeasworlds} follows this process to build world representations for quantitative physical reasoning. These studies show a shift from rule writing and state prediction toward 3D scene construction and physical inference, with video generation also used to combine executable models with visual realism.

\mbox{BWM-Forge} uses a Coding Agent to generate executable and editable Blender scenes and motion programs, renders videos with proxy models to define spatial relationships and motion paths, and then uses Seedance to add visual detail. Evaluation feedback guides further revisions. Compared with direct video generation, this approach keeps scenes and motion in code, reduces the need for the video model to infer them freely, and improves visual realism while retaining control over world states and motion.

\subsection{Robotics Simulation}

Coding Agents are also widely used in robot task planning and physical simulation. Code as Policies converts natural language instructions into executable robot control code, while VoxPoser generates 3D value maps through code and combines them with motion planning to control robot arms. GenSim \citep{wang2024gensim} and RoboGen \citep{wang2024robogen} use LLMs to generate robot simulation tasks, scenes, and training data; Eureka writes and optimizes reward functions to help robots learn complex motions. More recently, Code2Worlds-2 \citep{zhang2026code2worlds2} converts generated 3D scenes into executable MuJoCo environments for navigation, grasping, and object interaction, with validation based on collisions, contact, stability, and task success rates. Work also connects robot motion with visual generation: APOLLO Blender \citep{messina2025apollo} imports URDF robot models into Blender and generates joint animations. Robot-Factored World Models \citep{kim2026robotfactored} uses robot kinematics and explicit rendering to separate robot motion constraints from visual prediction of the environment, while RoboRender \citep{huang2026roborender} generates realistic robot manipulation videos from conditions such as simulated depth videos and robot masks.

Our Robotics Simulation work uses URDF models, motion data, and demonstration videos to provide structural and motion constraints. A Coding Agent constructs robot motions and object interactions, and video generation improves visual realism. The robot's defined joint structure, kinematic constraints, and task goals provide an editable basis for motion and support evaluation through joint errors, collisions, contact relationships, and task completion. This feedback then guides further improvements.

\subsection{Robot Training Data}

Embodied intelligence models need robot training data with diverse interactions to adapt to different objects, scenes, and viewpoints. However, collecting real-world demonstrations requires robot hardware and human operation, which makes continuous data expansion costly. DROID \citep{khazatsky2024droid} points out that the training data for many robot manipulation policies are still collected in a limited number of environments, with insufficient scene diversity. BridgeData V2 \citep{walke2023bridgedata} provides a large-scale manipulation dataset collected across 24 environments. However, all data are collected using the WidowX 250 robot arm, so the dataset does not cover different robot embodiments. RT-1 \citep{brohan2023rt1} uses about 130,000 real-robot demonstrations and covers more than 700 tasks. However, its public dataset provides images from only one camera view and does not include synchronized multi-view observations of the same interaction.

Synthetic data generation provides another way to expand robot training data, but there is still room for improvement in real-world transfer and visual quality. RoboGen \citep{wang2024robogen} uses generative models to automatically create tasks, scenes, and training supervision, allowing robot skill demonstrations to be generated continuously. However, its transfer to real environments is still limited by the simulation-to-real gap. ReBot \citep{fang2025rebot} combines simulated robot and object motions with inpainted real-world backgrounds. Its reported VBench visual quality score is 66.4\%, which is close to but still lower than the 70.1\% score of real robot videos.

To address the need for more diverse robot visual data, this work develops a video generation pipeline that supports different environments, robot embodiments, and observation viewpoints. The pipeline focuses on texture details, material and lighting quality, scene integration, and motion consistency in the generated videos. Current visual examples demonstrate the performance of the pipeline in these aspects. The pipeline can also incorporate external demonstrations, providing a practical way to increase both the scale and visual diversity of robot video data.

\section{Method}
\subsection{Preliminary and Task Definition}
Our goal is to generate high-fidelity videos of a specified robot performing a given task based on robot assets, a task description, and mode-dependent conditioning data. We define the system input as

\begin{equation}
\label{eq:system-input}
x = (m, r, u, c),
\end{equation}

where $m\in\{0,1,2\}$ denotes the input mode, $r$ denotes the robot assets, and $u$ denotes the natural language task description. The conditioning data $c$ depend on the input mode:

\begin{equation}
\label{eq:conditioning-data}
c=\begin{cases}I_0, & m=0,\\(I_0,D_{\mathrm{motion}}), & m=1,\\V_h, & m=2.\end{cases}
\end{equation}

Here, $I_0$ is the initial scene image, and $V_h$ is a human demonstration video. The structured motion data package is defined as $D_{\mathrm{motion}}=(A,s,f)$, where $A$ is a sequence of robot states, $s$ is the action space type, and $f$ is the trajectory sampling frequency.

The system first uses a coding agent to build an executable scene program. A 3D engine then executes the program and renders the resulting scene, producing the intermediate outputs:

\begin{equation}
\label{eq:scene-program}
P=\mathcal C_{\mathrm{prog}}(x),\qquad (S,B)=\mathcal R(P)
\end{equation}

Here, $\mathcal C_{\mathrm{prog}}$ denotes the scene construction process driven by the coding agent. The program $P$ specifies scene objects, spatial relationships, robot motion, and camera settings. The operator $\mathcal R$ denotes program execution and 3D rendering, while $S$ is the proxy scene model and $B$ is the proxy video.

The system then generates a candidate video from the input $x$ and the proxy video $B$:

\begin{equation}
\label{eq:candidate-video}
\hat V=\mathcal G(x,B),
\end{equation}

where $\mathcal G$ denotes the video generation process, and $\hat V$ is the candidate robot manipulation video to be evaluated. The final video $V_g$ is determined after quality evaluation, iterative refinement, and user confirmation.

The complete system workflow is expressed as

\begin{equation}
\label{eq:system-workflow}
\mathcal F(x)=(V_g,B,M)
\end{equation}

where $\mathcal F$ includes scene construction, video generation, result confirmation, and process logging. The process record $M$ contains execution reports, task status, stage summaries, and related metadata. These records support workflow recovery, result review, and reuse of prior experience.

\subsection{Harness System Overview}
\begin{figure}[!ht]
\centering
\includegraphics[width=.78\linewidth]{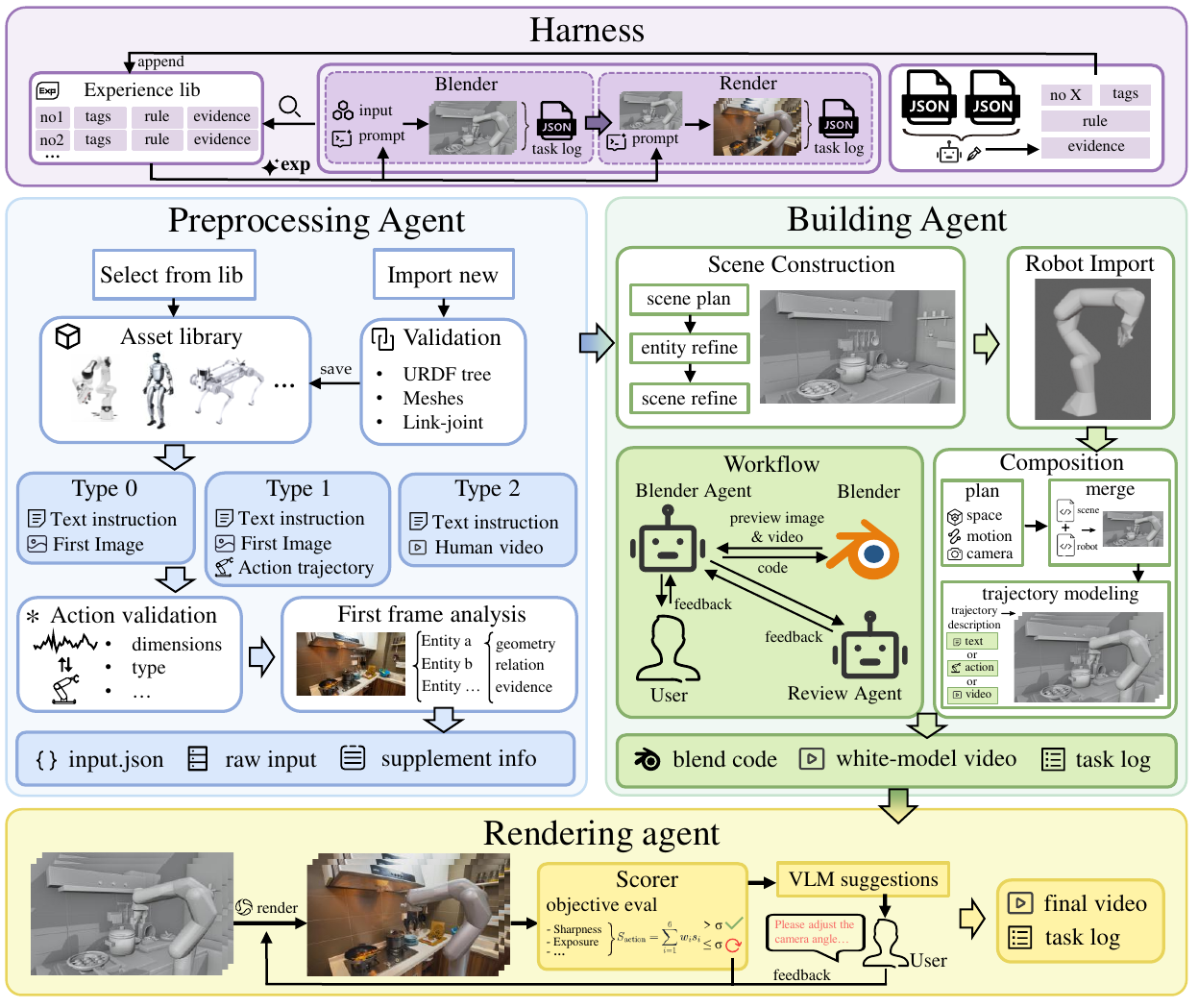}
\caption{Overview of \mbox{BWM-Forge}. The Harness coordinates input validation, scene construction, robot import, scene and robot composition, review, and video rendering. Task records and reusable experience connect the stages and support iterative refinement.}
\label{fig:method}
\end{figure}

We design a Harness system to manage the embodied data generation workflow driven by a coding agent. The workflow in Figure\nobreakspace{}\ref{fig:method} coordinates execution, information exchange, and feedback across four agent roles.

The Preprocessing Agent checks robot assets and motion data and extracts scene information from the initial scene image, or scene and motion information from a human demonstration video. It combines these information with the task description to prepare inputs for scene construction.

The Building Agent generates and executes Blender scripts to construct a 3D scene with the robot and its motion animation. It revises the scene based on user feedback.

After each modeling substage, the Review Agent checks scene layout, robot structure, and motion plausibility using visual evidence and execution logs. It makes local corrections when needed and records review results and unresolved issues.

The Rendering Agent uses the complete proxy video produced after all modeling stages are confirmed, together with the initial scene image and text prompts, to generate high-fidelity videos with a video generation model. Automated evaluation and user review guide video selection and iterative generation.

The Harness manages task states, intermediate outputs, execution logs, and stage summaries to support task state recovery and execution review. It also stores reusable modeling and video generation experience in a repository for future tasks.


\subsection{Preprocessing Agent}
Robot scene reconstruction must reconcile asset structures, motion constraints, and visual evidence. Input format inconsistencies or misalignment between motion data and robot assets can propagate through modeling. Because different error sources can produce similar discrepancies, validating only the final result complicates diagnosis and refinement. The Preprocessing Agent therefore progresses from basic data validation to task-specific interpretation.

Inputs comprise a first-frame image alone, a first-frame image with a motion trajectory, or a demonstration video. Across these modes, the Preprocessing Agent reuses three core modules: asset validation, motion consistency validation, and visual information parsing. Together, they construct a unified input context for the Building Agent's scene construction and scene and robot composition.

\paragraph{Asset Validation}

This step checks the format and completeness of robot assets, videos, motion trajectories, and first-frame images.

Robot asset validation covers the tree structure of the Unified Robot Description Format (URDF) model, mesh file references, associations between links and joints, and the numerical validity of poses. Video validation checks whether the container structure and video tracks can be parsed. Motion trajectory validation checks the file format (Parquet or NPZ) and the action space: joint space (Joint) or end-effector space (EEF). Detailed validation rules are provided in the appendix.

\paragraph{Motion Consistency Validation}

When $m=1$, this step verifies the mapping between $D_{\mathrm{motion}}$ and the target assets, together with the associated units and timing conventions.

The Preprocessing Agent parses the Parquet/NPZ motion data and checks data fields, array dimensions, frame counts, and numerical validity using the joint mapping, sampling rate, and initial state. It also verifies the correspondence between the motion data and the URDF joint structure, while establishing the initial values, units, and timing conventions. Detailed validation rules are provided in the appendix.

Asset validation establishes data validity, while motion consistency validation defines conventions for robot import and motion configuration. Inputs proceed to visual parsing and Building Agent execution only after asset validation and, for $m=1$, motion consistency validation pass.

\begin{figure}[!tp]
\centering
\includegraphics[width=1\linewidth]{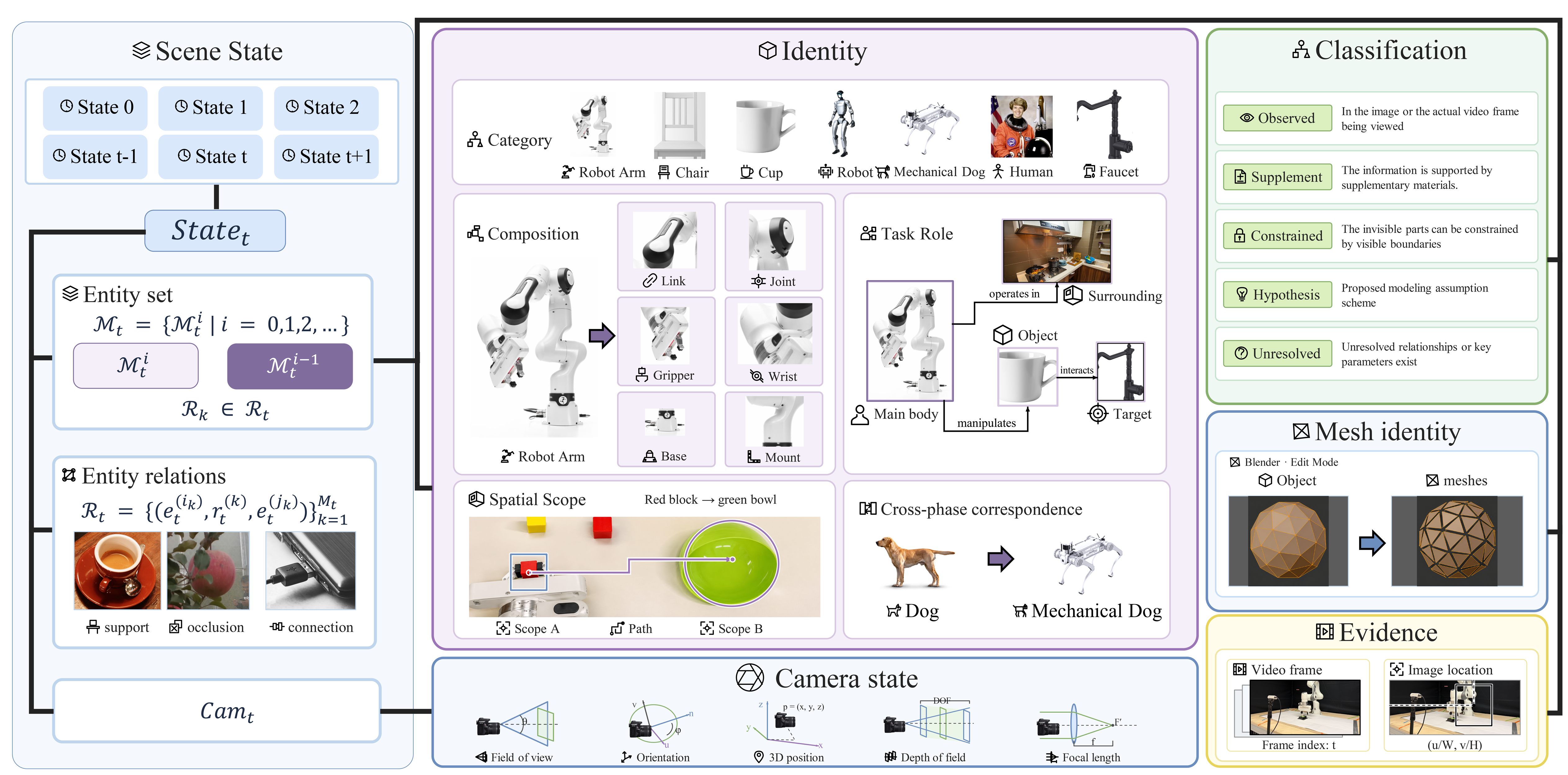}
\caption{Scene-state representation produced by the Preprocessing Agent. Frame-level states organize entity metadata, inter-entity relationships, and camera information. Entity identity, classification, mesh associations, and visual evidence support subsequent scene construction.}
\label{fig:process-agent}
\end{figure}

\paragraph{Visual Information Parsing}

Visual parsing produces a semantic description of the scene state $State_{4D}$ from the raw image $I_0$ or video $V_h$, organized around entity metadata, camera information, and inter-entity relationships.

To support both scene understanding and object-level modeling, we organize each frame's state $State_t$ into entity states $\mathcal{M}^{i}_t$, inter-entity relationships $\mathcal{R}_t$, and the camera state $Camera_t$.

This representation decomposes visual information into independently accessible and editable entity-level modeling evidence. It allows the Building Agent to select objects, initialize geometry, and associate models using entity metadata, and to determine the scene layout and check spatial constraints using inter-entity relationships.

To construct this representation from visual inputs, we first preprocess $I_0$ or $V_h$ to obtain a frame set $\mathcal{F}_{\mathrm{vis}}$. For $m\in\{0,1\}$, $\mathcal{F}_{\mathrm{vis}}=\{I_0\}$. The Coding Agent then performs semantic scene parsing on each frame $F_t$ in $\mathcal{F}_{\mathrm{vis}}$. It identifies the main entities' geometric components and initial states and extracts their support, occlusion, and relative spatial relationships to produce a conceptual description of the corresponding scene state $\mathrm{State}_t$:

\begin{equation}
\begin{aligned}
&\mathrm{State}_{4D}
  = \{\mathrm{State}_t\}_{t\in\mathcal{V}_{\mathrm{vis}}},\\
&\mathrm{State}_t
  = \left\{\{\mathcal{M}_t^{(i)}\}_{i=1}^{N_t},
      \mathrm{Camera}_t,\mathcal{R}_t\right\},\\
&\mathcal{R}_t
  = \left\{
      \left(e_t^{(i_k)},r_t^{(k)},e_t^{(j_k)}\right)
    \right\}_{k=1}^{M_t}.
\end{aligned}
\end{equation}

Here, $\mathcal{R}_t$ is the set of identified relationships between entities in the scene. $e_t^{(i_k)}$ and $e_t^{(j_k)}$ are the two entities involved in the $k$-th relationship, and $r_t^{(k)}$ denotes its type, such as support, occlusion, or another relative spatial relationship. $\mathcal{M}_{t}^{(i)}$ is the metadata for entity $i$ at time $t$.

The entity-level metadata $\mathcal{M}_{t}^{(i)}$ organizes entity-related information into fields such as \texttt{identity}, \texttt{classification}, \texttt{mesh\_\allowbreak{}identity}, \texttt{evidence}, \texttt{completion}, and \texttt{unknown\_\allowbreak{}fields}. These fields serve as semantic dimensions for analyzing and organizing information. Reference structures for $\mathcal{M}_{t}^{(i)}$, $State_{4D}$, and $\mathcal{R}_t$ are shown in Figure\nobreakspace{}\ref{fig:process-agent}.

For scene objects whose absolute scale cannot be determined directly because calibration information is unavailable, dimensions and spatial layouts are represented using relative proportions or editable parameters. During subsequent scene construction, the known dimensions of the target model are used for scale alignment and scene registration, reducing the effect of visual estimation errors on the geometric consistency of the scene.

After asset validation, motion validation, and first-frame information extraction, the extracted first frame and associated information are made available as modeling evidence for the Building Agent. This gives the Building Agent access to the target model, visual references, and motion conventions.

Let $k=1,2,3,\ldots$ denote successive modeling iterations. The conversational context $\mathcal{C}_{\mathrm{build}}^{(k)}$ for iteration $k$ is defined as

\begin{equation}
\label{eq:building-context}
\begin{aligned}
\mathcal{C}_{\mathrm{build}}^{(k)}
= \operatorname{Assemble}\bigl(&\Pi,x,\mathcal{F}_{\mathrm{vis}}^{(k)},S^{(k)},K^{(k)},\\
&F^{(k-1)},H^{(k)},V_r^{(k-1)},\mathrm{State}_{4D}\bigr).
\end{aligned}
\end{equation}

The prompt $\Pi$ comprises the user prompt $\Pi_{\mathrm{user}}$, a general modeling prompt $\Pi_{\mathrm{gen}}$ shared across input modes, and a prompt $\Pi_{m,stage}$ specific to input mode $m$ and modeling stage $stage\in\{SceneConstruction,\allowbreak Robot\ Import,\allowbreak Composed\}$. $\mathcal{F}_{\mathrm{vis}}^{(k)}$ is the keyframe set for the current modeling iteration, $x$ is the original system input, and $K^{(k)}$ is the experience retrieved for the current iteration. Supplementary information $S^{(k)}$ comprises textual information $S_{\mathrm{text}}^{(k)}$ and image information $S_{\mathrm{img}}^{(k)}$. $H^{(k)}$ contains the shared conversation and tool execution records from previous iterations; $\mathrm{State}_{4D}$ is the scene state obtained through visual parsing; and $V_r^{(k-1)}$ contains the project and preview results from the previous iteration. $F^{(k-1)}$ denotes feedback from the user or the Review Agent on the preceding modeling result. $\operatorname{Assemble}$ organizes text instructions, images, and accessible file references into the contexts of the Building Agent and the Review Agent.

\FloatBarrier

\subsection{Building Agent}
\begingroup
\setlength{\parskip}{6pt plus 1pt minus .5pt}
\widowpenalty=10000
\clubpenalty=10000
\newcommand{\buildingphase}[1]{\par\noindent{\normalfont\normalsize\bfseries #1}\enspace\ignorespaces}
For inputs $x$, conditioning data $c$, and input modes $m$, embodied video generation requires consistency among the visual scene, robot structure, and motion. Scene geometry is inferred from images or videos, while robot geometry and joint constraints follow the target assets. The robot's scale, position, and trajectory must then be aligned with the scene. Handling these tasks in one stage can cause interference and obscure the source of modeling errors.

\begin{figure}[!tp]
\centering
\includegraphics[width=1\linewidth]{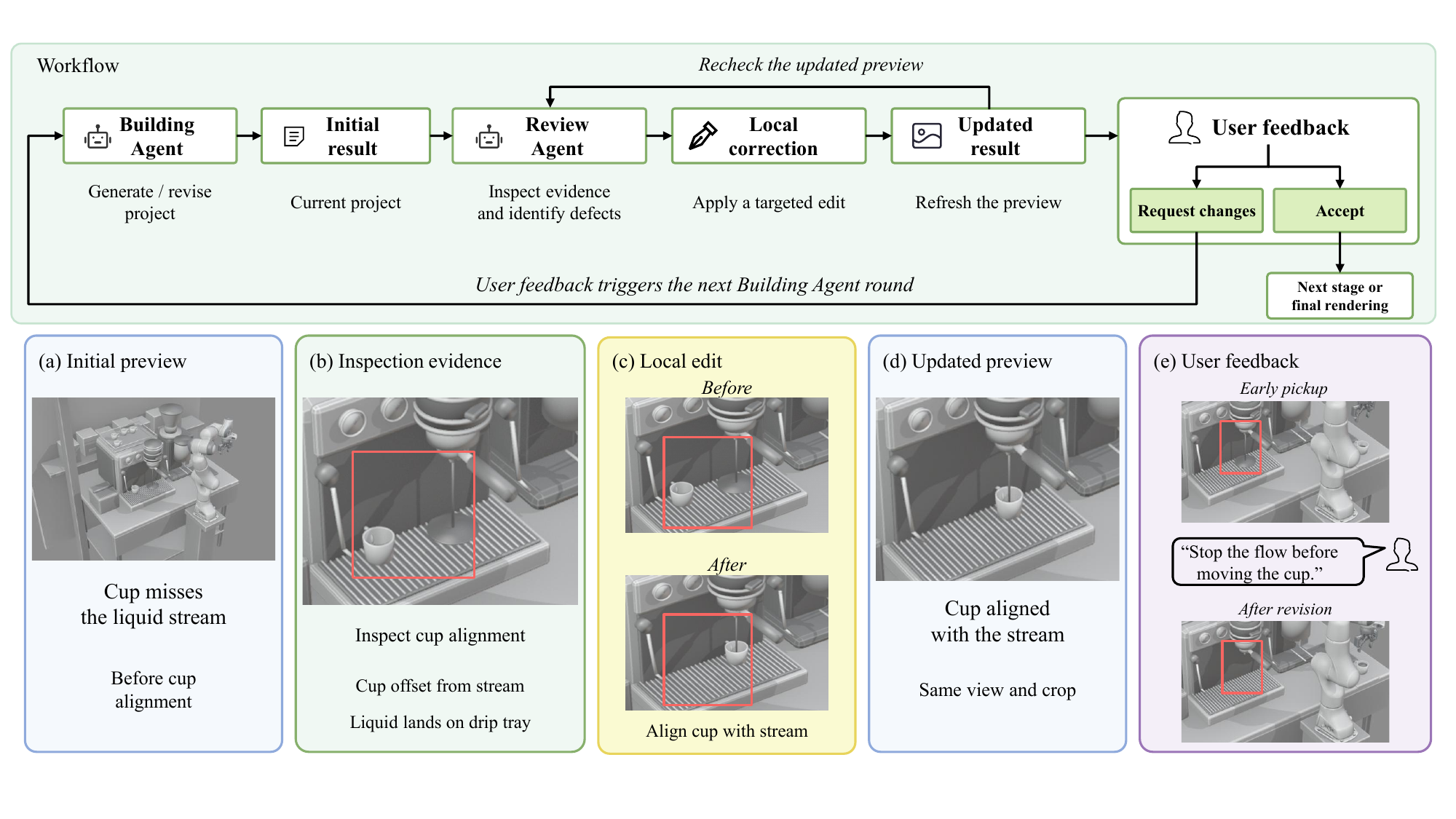}
\caption{Building and review workflow. The Review Agent inspects the current project and identifies defects, which guide local corrections. The updated preview is checked again before user review. User feedback can trigger another Building Agent round, while acceptance advances to the next stage or final rendering. The example illustrates a local adjustment to the placement of foreground objects.}
\label{fig:building-review}
\end{figure}

We therefore adopt a prompt-guided, coarse-to-fine strategy with three stages. Scene Construction recovers the environment layout and task-relevant entities. Robot Import imports the robot structure and geometry. And Composition integrates both through spatial registration, animation construction, and constraint validation. Separating these stages supports targeted refinement from global structure to local geometry. After modeling, the Building Agent summarizes the modeling process and submits candidate experience entries to the experience repository.

\subsubsection{Scene Construction}

The scene construction stage uses the first-frame image $I_0$ and the state information $State_{4D}$ extracted by the Preprocessing Agent as its primary evidence. It progressively reconstructs the task-relevant environment and reference camera, proceeding from the global layout to local geometry, and produces a standalone scene modeling project.

Scene planning begins with camera cues. The Preprocessing Agent's camera parsing results describe the viewpoint, perspective, and framing cues in the reference image at a semantic level. The Building Agent establishes and adjusts $Camera_t$ using these cues.

The overall scene layout and inter-entity relationships are initialized from the viewpoint of $Camera_t$. Each entity is designated for retention, replacement, exclusion, or further confirmation, and task objects and their relationships are identified using $\mathcal{R}_t$. Initial geometric representations of the main entities are then established, prioritizing a plausible overall scene layout and task relationships.

Based on this layout, a list of entities to model is constructed from the entity metadata extracted by the Preprocessing Agent. For each entity, $\mathcal{M}^{(i)}_t$ specifies its identity, category, candidate mesh assets, visual evidence, and information completeness. The entity's designated status from layout planning is then used to determine an appropriate modeling strategy.

The geometry and scene representations of key entities are progressively refined, including their overall shapes, relative positions, orientations, and initial states, as well as support, occlusion, and depth ordering between entities. The corresponding Blender objects and meshes are organized by entity identity using the \texttt{identity} field in the entity metadata, and the associations between entity semantics and objects in the modeling project are recorded. For objects without reliable scale information, dimensions are expressed as relative proportions or editable parameters, with the basis for inferred values, completed geometry, and unresolved parameters documented.

The scene is then optimized by comparing images rendered from the reference camera with the visual reference, considering both the overall composition and entity-level visual details. These comparisons check the positions, silhouettes, and occlusion relationships of the main entities. Metrics such as intersection over union (IoU) and mean intersection over union (mIoU) provide auxiliary evidence for these checks. If a check fails, the scene entities are iteratively remodeled and refined.

Scene Construction outputs a standalone scene project and its rendered results, records unresolved geometry and camera parameters, and submits them to the user or the Review Agent for review. Feedback from the user or the Review Agent can trigger further corrections to geometry or layout, closing the refinement loop for scene construction.

\subsubsection{Robot Import}

A standalone robot model is created from the input robot assets $r$, preserving the target robot's complete URDF structure, visual geometry, transformations, and scale. Preview images are prepared and submitted to the user or the Review Agent for iterative adjustment.

\subsubsection{Composition}

The scene and robot composition stage combines the standalone projects from Scene Construction and Robot Import into a shared scene. It jointly models the robot and environment and generates a 3D proxy video, progressing from global spatial registration to local motion refinement and then validation of dynamic constraints. The system first plans spatial alignment, motion, and keyframes for validation; it then models motion and animation, followed by validation of dynamic constraints and iterative correction.

\buildingphase{Planning.} This stage covers the spatial configuration of the robot, motion, camera control, and validation.

The target robot is registered in the scene exported by Scene Construction. Its relative scale, overall position, orientation, and initial joint configuration are established to satisfy the intended composition and task relationships.

For motion planning, the system uses the input motion trajectory or demonstration video together with the task description to determine the range of motion, timing conventions, and motion sequence to be reconstructed.

The system extracts action-related instructions $\Pi_{\mathrm{act}}$ from the user prompt $\Pi_{\mathrm{user}}$. Motion planning follows the input mode:
\par\smallskip
\begingroup
\setlength{\parskip}{0pt}
\setlength{\tabcolsep}{7pt}
\renewcommand{\arraystretch}{1.0}
\noindent\begin{tabularx}{\linewidth}{@{}>{\raggedright\arraybackslash}p{.10\linewidth}>{\raggedright\arraybackslash}X@{}}
\toprule
{\normalfont\bfseries Mode} & {\normalfont\bfseries Motion planning} \\
\midrule
$m=0$ & Where no explicit trajectory is available, the required motion is inferred from $\Pi_{\mathrm{act}}$. \\
\addlinespace[4pt]
$m=1$ & The input trajectory serves as a motion reference. Limited retargeting or correction is performed based on first-frame correspondences, validity under the URDF specification, and task constraints to compensate for scene modeling errors. \\
\addlinespace[4pt]
$m=2$ & The required motion and timing conventions are inferred from the video $V_h$, entity states $\mathcal{M}^{i}_t$, and $\Pi_{\mathrm{act}}$. \\
\bottomrule
\end{tabularx}
\par\endgroup\smallskip

For camera control, the system extracts camera-related instructions $\Pi_{\mathrm{cam}}$ from the user prompt $\Pi_{\mathrm{user}}$ and translates the camera motion intent expressed in natural language into an executable camera control plan. It first parses the camera motion type, direction, target, viewpoint changes, speed, and time interval to determine the overall motion pattern and framing objectives. It then generates viewpoint, perspective, and framing information for the keyframe cameras $\mathcal{C}_{\mathrm{ref}}$ and combines this information with the intended camera motion to generate or adjust camera control code:

\begin{equation}
\label{eq:camera-planning}
\begin{aligned}
\widehat{C}_t &= \operatorname{Plan}_{\mathrm{cam}}\left(\mathcal{C}_{\mathrm{ref}},\Pi_{\mathrm{cam}}\right),\\
\mathcal{C}_{\mathrm{ref}} &= \left\{\mathrm{Camera}_t\mid F_t\in\mathcal{F}_{\mathrm{vis}}\right\}.
\end{aligned}
\end{equation}

Here, $\widehat C_t$ is the completed camera state at time $t$; $\mathcal{C}_{\mathrm{ref}}$ is the collection of viewpoint, perspective, and framing information for the keyframe cameras; and $\Pi_{\mathrm{cam}}$ denotes camera motion planning constraints parsed from the text prompt. $\operatorname{Plan}_{\mathrm{cam}}$ generates control code for intermediate camera states using the reference camera states and motion constraints.

The planning procedure $\operatorname{Plan}_{\mathrm{cam}}$ considers the consistency of keyframe framing, the temporal continuity of camera motion, and target visibility to reduce abrupt trajectory changes and viewpoint drift.

For validation, the Building Agent selects frames or samples additional frames as needed to form $\mathcal{F}_{rev}$, based on task events, the current animation, and high-risk locations. The keyframe set $\mathcal{F}_{rev}$ is expanded during motion planning $\operatorname{Plan}_{\mathrm{act}}$  and camera planning $\operatorname{Plan}_{\mathrm{cam}}$. The Composition stage and the Review Agent validate the planned keyframes in both $\mathcal{F}_{\mathrm{vis}}$ and $\mathcal{F}_{\mathrm{rev}}$ and provide feedback.

\buildingphase{Motion and animation modeling.} Guided by the planning results, the system merges the projects, refines geometric relationships, and performs detailed modeling.

The system first merges the Blender projects generated during Scene Construction and Robot Import, preserving scene objects, the robot's full structural hierarchy, visual meshes, and original scale information. It also unifies coordinate transformations and motion scaling.

Building on the spatial registration results, the system further adjusts the dimensions, positions, orientations, and initial poses of relevant entities. The system compares rendered images of the composed scene with the original first frame to identify visual discrepancies, progressively refining the result from overall spatial alignment to correction of local geometric relationships.

The camera trajectory and control parameters are further refined based on checks of target visibility, image composition, and motion continuity at the keyframes in $\mathcal{F}_{\mathrm{vis}}$.

Motion modeling addresses root motion, including the robot's overall translation, turning, and body state, as well as changes in individual joints. For motions that require inference, joint targets are determined from the robot structure, initial state, and task constraints, with checks for joint limits and, where necessary, reachability. Animation modeling covers background entities and entities that move in coordination with the robot. During modeling, the keyframes in $\mathcal{F}_{\mathrm{rev}}$ are determined around motion onset and completion, state transitions, motion extrema, and high-risk positions.

\buildingphase{Motion refinement.} After modeling, the system checks physical constraints at the keyframes in $\mathcal{F}_{\mathrm{vis}}$, renders a proxy video, and submits it to the Review Agent. Failed checks trigger iterative correction.

At keyframes and necessary intermediate times, the system checks motion continuity, contact relationships, sliding, support stability, obstacle interference, and potential collisions in the space swept by the robot. When it detects a problem, it identifies the relevant time interval, robot components, and scene objects, then makes targeted adjustments to geometric parameters, spatial poses, or motion trajectories. The animation is regenerated for validation.

Once checks at the keyframes in $\mathcal{F}_{\mathrm{vis}}$ and the necessary continuous-motion checks meet predefined requirements, the system renders the complete proxy video $B$ and submits it to the user or the Review Agent for review, yielding feedback $F^{k}_{building,composed}$.

User feedback can trigger further corrections to spatial registration, local geometry, or motion until the result satisfies the task requirements. Further details of the checks are provided in the appendix.

\subsubsection{Candidate Experience Summarization}

After Scene Construction, Robot Import, and Composition pass review, references to evidence such as stages, iterations, event records, images, and scripts are supplied as text excerpts. Coding Agent uses these excerpts to summarize candidate experience entries. The Harness proactively retrieves candidate experience entries prepared by the Building Agent and Rendering Agent, decides how to update the experience repository based on their scope of applicability and evidence sources, and updates the local repository. The implementation is described in detail in Experience Repository Construction and Evolution.

\FloatBarrier
\endgroup

\subsection{Review Agent}
The Review Agent combines visual and numerical evidence from execution results with the task objectives to determine whether the current proxy animation and rendered video are valid. When a result fails, it locates the problem and performs targeted repairs (Figure\nobreakspace{}\ref{fig:building-review}). Human feedback is optional: the Review Agent approximates human judgments of the proxy scene previews and generated videos, then submits the results to the user for feedback.

For visual review, the keyframes $\mathcal{F}_{\mathrm{vis}}$ and randomly sampled frames $\mathcal{F}_{\mathrm{random}}$ form a frame set $\mathcal{F}$. The corresponding frames of the proxy video are reexamined with reference to the general prompt $\Pi_{gen}$, user prompt $\Pi_{{User}}$, and user feedback $F^{(k-1)}$. This image-level review checks whether the outputs of Scene Construction, Robot Import, and Composition are consistent with these references.

When visual evidence is inconsistent with the prompts, the Review Agent revisits the numerical evidence $H_{data}$ in execution logs and recorded motion trajectories, and records the trajectory values associated with the defects.

Based on these checks, the Review Agent makes local corrections to clearly identified, localized problems arising in Scene Construction, Robot Import, or Composition. It records any remaining issues and submits them to the user for feedback.

\subsection{Rendering Agent}
The Rendering Agent first retrieves relevant experience from the experience library and incorporates it into the prompt. It then renders and refines the video, outputs the resulting video $V$, and produces stage documentation.

\paragraph{Objective Evaluation}

We introduce seven objective scoring metrics to evaluate the quality of the generated video:

\begin{itemize}

\item \textbf{Sharpness:} Measures the clarity of image details and edges.

\item \textbf{Exposure:} Assesses whether the video is too dark or too bright.

\item \textbf{Color stability:} Measures color consistency across frames.

\item \textbf{Temporal flicker:} Detects sudden changes in brightness over time.

\item \textbf{Motion smoothness:} Assesses whether motion is smooth and continuous.

\item \textbf{Scene stability:} Measures the consistency of the surrounding scene across frames.

\item \textbf{Action fidelity:} Assesses how closely the generated motion matches the reference video.

\end{itemize}

Sharpness and exposure assess spatial quality. Color stability, temporal flicker, motion smoothness, and scene stability assess temporal consistency. Action fidelity evaluates consistency with the reference motion.

Together, these metrics evaluate visual detail, brightness, color consistency, motion continuity, action consistency, and scene stability. Action fidelity is computed as a weighted sum of six component scores:

\begin{equation}
\label{eq:action-fidelity}
\begin{aligned}
&S_{\mathrm{action}} = \sum_{i=1}^{6} w_i s_i,\\
&\mathbf{w} = (0.25,\;0.15,\;0.35,\;0.10,\;0.10,\;0.05)^{\mathsf{T}}.
\end{aligned}
\end{equation}

Here, $s_1$ through $s_6$ denote motion curve correlation, motion intensity similarity, direction similarity, activity interval overlap, start and end timing similarity, and duration similarity, respectively. The vector $\mathbf{w}=(w_1,\ldots,w_6)^{\mathsf{T}}$ contains the corresponding nonnegative weights, which sum to one.

The objective evaluation produces individual metric scores and a PASS / RETRY / REJECT decision. The Rendering Agent then identifies the three most critical issues based on the metric weights and score deficits relative to a predefined acceptance threshold $\tau$. These issues guide the automatic correction stage.

\begin{figure}[!tp]
\centering
\includegraphics[width=0.8\linewidth]{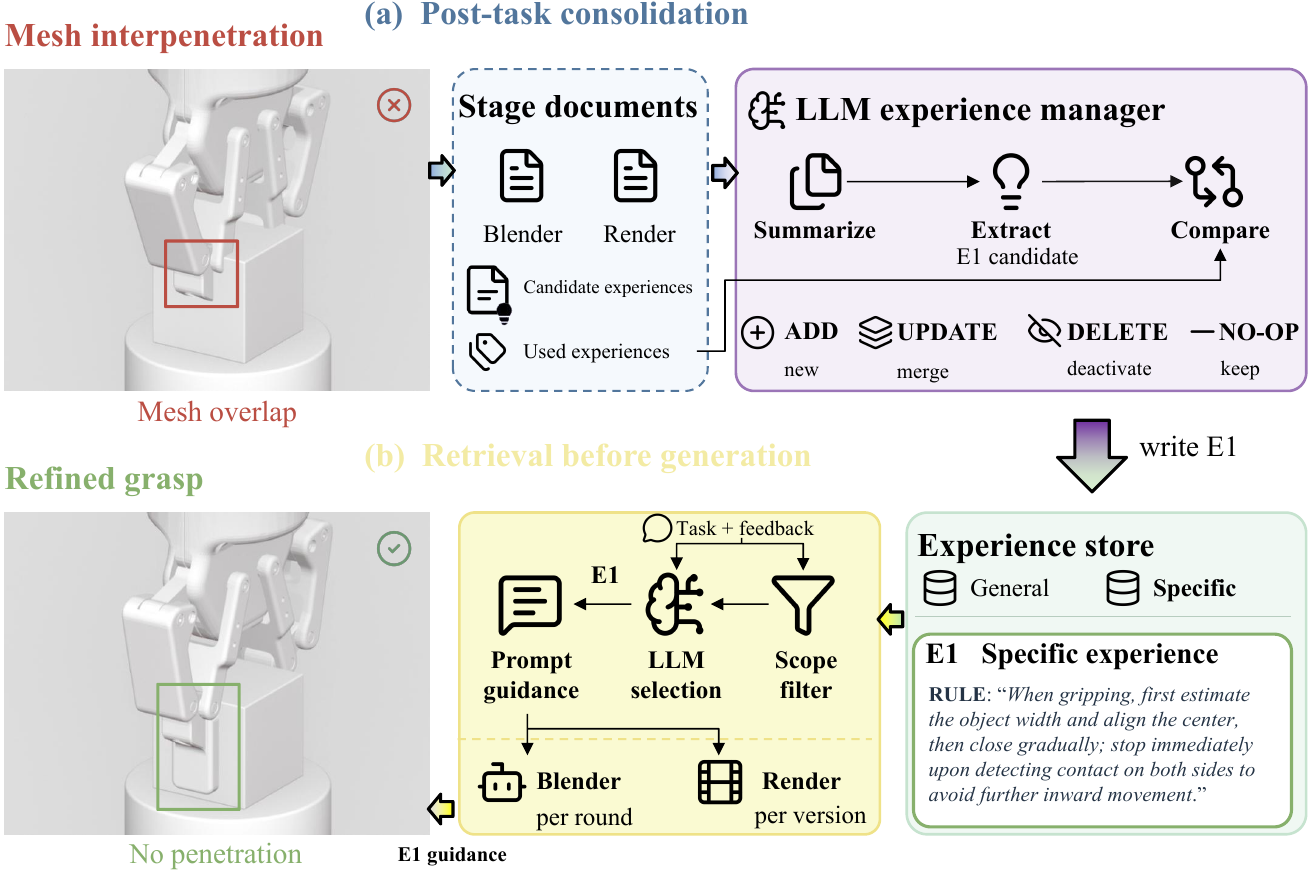}
\caption{Experience-library update and retrieval. Stage documents provide candidate rules for creation, update, deactivation, or no-op decisions. Before generation, scope filtering and LLM selection retrieve applicable experience to guide the Building and Rendering Agents; the illustrated grasping rule addresses mesh interpenetration.}
\label{fig:experience-library}
\end{figure}

\paragraph{Automatic Correction: Test-Time Prompt Optimization}

For an initial video with an objective evaluation score below $\tau$, the Rendering Agent renders it again. Based on the problems identified in the scoring feedback, the agent adds the corresponding correction instructions to the rendering prompt. It then generates one automatically corrected video.

If the corrected video still scores below $\tau$, it proceeds to human review without further automatic correction. Initial and corrected videos that meet or exceed $\tau$ proceed directly to visual evaluation by a vision-language model.

\paragraph{Vision-Language Model Evaluation}

We introduce visual evaluation using a vision-language model (VLM) to better assess the visual quality of the generated video.  The evaluation covers visual realism, physical and temporal coherence, and task semantic consistency. The detailed scoring criteria and weights are provided in Appendix\nobreakspace{}\ref{app:review}.

The VLM receives the compressed reference first frame and chronological frame montages from the proxy video and the rendered video. Each montage contains eight frames by default. It also receives 12 additional sets of action comparison frames from the two videos, the current prompt, the seven objective scores, and color shift statistics relative to the reference image.

While scoring the video, the VLM also checks the actions in three stages: initial action, grasping, and carrying and placement. For each stage, it returns a PASS / FAIL / UNCERTAIN result, timestamps from both videos, and visible evidence. These scores and evaluations serve as references for human review.

Let $S_{\mathrm{obj}}$ denote the objective score and $S_{\mathrm{VLM}}$ denote the VLM visual score. The final video score is computed as a weighted sum:

\begin{equation}
\label{eq:video-score}
S = \lambda S_{\mathrm{obj}} + (1-\lambda)S_{\mathrm{VLM}},
\end{equation}

where $\lambda$ denotes the weight assigned to the objective score.

After scoring, the user can preview the generated video and its scores and decide whether to modify it. For each revision round, the rendering inputs include the previous video's scores, VLM evaluation, and user revision comments. Revision rounds do not include automatic correction.

After the user confirms the rendering result, the agent summarizes the prompts, scores, and user feedback from the entire stage. It generates a stage summary document and adds it to the experience library.

\subsection{Experience Repository Construction and Evolution}
The experience library in Figure\nobreakspace{}\ref{fig:experience-library} stores experience summaries and archives. It consists of general experience and specialized experience. General experience includes rules and methods that apply across tasks or objects. Specialized experience includes rules for specific robots, grippers, task types, scenes, or assets.

The Building Agent and Rendering Agent search the experience library at the start of each task. They construct a query from the task description, robot, scene, and current feedback, then use a large language model (LLM) to retrieve relevant experience. The system first excludes inactive entries and entries whose scope conflicts with the current task. The remaining entries are provided to the LLM for selection. Reference information from the selected entries is then added to the corresponding task prompts.

At the end of each stage, the Building Agent and Rendering Agent use the LLM to summarize the stage conversations and results. The recorded fields differ between the two stages. The Building stage mainly records the status of each modeling stage and revision round, user feedback, model files, preview images, event logs, check results, and experience used. The Render stage mainly records the prompts for each video version, user revision requests, generation status, evaluation results, scores, problem lists, revision decisions, and experience used.

Experience entries are extracted from the summary documents produced by both stages. These documents include candidate entries that describe potentially reusable rules. The experience library uses the LLM to review these candidates and extract additional experience from the stage documents. The LLM determines whether each candidate is suitable for inclusion in the library.

For each selected candidate, the LLM compares its rules with the existing library content and proposes one of the following operations:

\begin{itemize}

\item \textbf{Create:} Add a new experience entry.

\item \textbf{Update:} Merge the candidate into an existing experience entry.

\item \textbf{Delete:} Deactivate an existing experience entry.

\item \textbf{No-op:} Ignore the current candidate.

\end{itemize}

This process connects the Building Agent, Rendering Agent, and experience library through a feedback loop. Agent execution produces stage summaries, which guide updates to the library. The accumulated experience supports subsequent tasks and helps improve execution efficiency and generation quality over time.

\FloatBarrier

\begingroup
\setlength{\parskip}{4pt plus 1pt minus .5pt}
\setlength{\intextsep}{8pt plus 2pt minus 2pt}
\titlespacing*{\subsubsection}{0pt}{8pt}{4pt}
\section{Experiments}
\subsection{Experimental Setup}
\subsubsection{Dataset and Implementation Details}

We selected 15 examples from the three input modes supported by \mbox{BWM-Forge}. They cover five robot types: a single arm, dual arms with grippers, dual arms with dexterous hands, a quadruped, and a humanoid robot. The scenes cover everyday living, office, industrial, and laboratory environments, as well as surgical training settings, among others. We retained each original \mbox{BWM-Forge} video and created a corresponding 16 FPS version. LingBot-World generated a comparison video for each task. We evaluated all three groups with VBench-Long and RoboJudge-9B.

The Input agent uses gpt-5.6-sol to parse user requests and imported asset packages. The Blender agent uses gpt-6-astra to guide scene construction and proxy rendering. The Blender stage Harness summary agent uses gpt-5.6-sol. The Render stage uses gpt-5.6-sol for visual evaluation, feedback processing, and Harness summaries. We use Blender 5.2.2 LTS to build the proxy scenes and render their videos. The final appearance videos are generated by Doubao-Seedance-2.5.

The baseline uses the official lingbot-world-base-cam weights. Each task receives one reference image and an explicit prompt. The original tasks use their corresponding reference images. The dexterous hand task uses the first frame of the final \mbox{BWM-Forge} video, so its baseline input includes appearance information generated by \mbox{BWM-Forge}. Video resolutions, frame rate conversion details, and baseline settings are provided in Appendix C.

\subsubsection{Evaluation Metrics}

We use six VBench-Long metrics to assess video quality. Aesthetic Quality (AQ) measures the visual appeal of individual frames, including composition, color richness and harmony, photographic realism, natural appearance, and artistic quality. Imaging Quality (IQ) assesses image distortions such as overexposure, noise, and blur. Subject Consistency (SC) measures whether the subject's appearance remains consistent across frames. Background Consistency (BC) measures the consistency of the background over time. Motion Smoothness (MS) assesses how smooth and visually realistic the motion is. Dynamic Degree (DD) measures the presence of large movements. A higher DD indicates more dynamic content, rather than greater task success.

We use two RoboJudge-9B metrics to assess embodied task videos. Physical Adherence (PA) measures physical plausibility and visual consistency without considering the task instruction. It covers robot structure and temporal consistency, scene and object consistency, and the realism of contact, grasping, object motion, and responses to forces.

Instruction Alignment (IA) measures whether the video follows the text instruction and initial image to complete the task. It considers robot identity, the target object and its properties, and the final goal state. Both PA and IA use a scale from 1 to 5. Higher scores indicate better performance on the corresponding metric.

\subsection{Quantitative Results}
\subsubsection{VBench-Long Results}

Table\nobreakspace{}\ref{tab:vbench} reports equally weighted means over the 15 source videos on the original VBench aggregate scale. \mbox{BWM-Forge} scores higher than LingBot-World Base Cam in aesthetic quality, subject consistency, background consistency, and motion smoothness. The largest gap among these four metrics is in subject consistency, with scores of 0.972146 and 0.947975, respectively. Background consistency and motion smoothness show the same trend. These results indicate more stable subject and background appearance and smoother motion in the \mbox{BWM-Forge} videos for the selected tasks. Both versions of \mbox{BWM-Forge} score lower in dynamic degree than \mbox{LingBot-World}, indicating less large-scale motion. In imaging quality, the original videos score slightly lower, while the 16 FPS version scores marginally higher. Both versions score higher than \mbox{LingBot-World} in aesthetic quality, subject consistency, background consistency, and motion smoothness.

After conversion to 16 FPS and re-encoding, \mbox{BWM-Forge}'s imaging quality increases slightly from 0.732059 to 0.734156. Motion smoothness decreases from 0.994212 to 0.991300, while subject and background consistency each change by less than 0.001. The relative ordering against LingBot-World remains unchanged for five metrics. Imaging quality becomes marginally higher than LingBot-World’s score of 0.734126, with a difference of only 0.000030. Thus, \mbox{BWM-Forge} retains its advantages in aesthetic quality, consistency, and motion smoothness after the 16 FPS conversion on these 15 examples.

\begin{table}[htbp]
\centering
\caption{Six VBench-Long metrics for 15 tasks.}
\label{tab:vbench}
\begingroup
\small
\setlength{\tabcolsep}{4pt}
\renewcommand{\arraystretch}{1.15}
\begin{tabularx}{\linewidth}{@{}Xrrrrrrr@{}}
\toprule
\textbf{Method and video version} & \textbf{Videos} & \textbf{AQ} & \textbf{IQ} & \textbf{SC} & \textbf{BC} & \textbf{MS} & \textbf{DD} \\
\midrule
LingBot-World Base Cam & 15 & 0.513536 & 0.734126 & 0.947975 & 0.957831 & 0.987241 & 0.788889 \\
\mbox{BWM-Forge} & 15 & 0.522603 & 0.732059 & 0.972146 & 0.972240 & 0.994212 & 0.502319 \\
\mbox{BWM-Forge} 16 FPS & 15 & 0.528396 & 0.734156 & 0.971763 & 0.972115 & 0.991300 & 0.511247 \\
\bottomrule
\end{tabularx}
\endgroup
\end{table}

\subsubsection{RoboJudge-9B Results}

Table\nobreakspace{}\ref{tab:robojudge} reports the mean RoboJudge-9B scores over the 15 tasks. The original \mbox{BWM-Forge} videos obtain PA and IA scores of 2.93 and 3.33, compared with 2.89 and 2.73 for \mbox{LingBot-World Base Cam}. The difference is larger in IA, where \mbox{BWM-Forge} scores 0.60 points higher. This result indicates closer alignment with the input instructions and target states on the selected tasks. \mbox{BWM-Forge} also receives a slightly higher PA score than \mbox{LingBot-World}, with a difference of 0.04 points. This gap is smaller than its 0.60-point advantage in instruction alignment.

After conversion to 16 FPS and re-encoding, \mbox{BWM-Forge}'s PA score increases from 2.93 to 3.00, while its IA score increases from 3.33 to 3.40. These changes reflect the combined effects of frame rate adjustment and re-encoding.

The other models in Table\nobreakspace{}\ref{tab:robojudge} are evaluated using official human scores on different tasks, videos, and sample sizes. These values provide a reference for the score range and are not used to rank models against automatic scores in this study.

\begin{table}[htbp]
\centering
\caption{RoboJudge-9B automatic scores in this study and official human scores for reference.}
\label{tab:robojudge}
\begingroup
\small
\setlength{\tabcolsep}{4pt}
\renewcommand{\arraystretch}{1.15}
\begin{tabularx}{\linewidth}{@{}Xrrr@{}}
\toprule
\textbf{Method or generation model} & \textbf{Videos} & \textbf{PA} & \textbf{IA} \\
\midrule
\mbox{BWM-Forge} & 15 & 2.93 & 3.33 \\
\mbox{BWM-Forge} 16 FPS & 15 & 3.00 & 3.40 \\
LingBot-World Base Cam & 15 & 2.89 & 2.73 \\
abot\_physworld\_f113 & 94 & 3.45 & 3.57 \\
cogvideo\_i2v\_f113 & 96 & 1.79 & 2.09 \\
cosmos3 & 107 & 3.44 & 3.61 \\
gigaworld\_gr1\_2b\_f113 & 105 & 2.74 & 3.03 \\
happyhorse & 36 & 3.81 & 4.33 \\
kling & 26 & 3.88 & 4.50 \\
veo31\_lite & 35 & 3.89 & 3.97 \\
wan22\_ti2v5b\_1280x704\_f169 & 110 & 2.60 & 2.63 \\
wan26\_flash & 69 & 4.09 & 4.16 \\
wow\_wan14b\_f105 & 122 & 2.84 & 2.74 \\
\bottomrule
\end{tabularx}
\endgroup
\end{table}

\FloatBarrier
\endgroup
\subsection{Qualitative Results}
For each example, we uniformly sample 30 frames from the proxy and final rendered videos. Figures\nobreakspace{}\ref{fig:example-01} to\nobreakspace{}\ref{fig:example-07} compare the scene and motion references with the resulting appearance across seven tasks and five robot types.

The manipulation examples emphasize interactions with nearby task objects, whereas the locomotion examples expose the relation between robot pose, scene layout, and camera motion. In both groups, the proxy model makes the intended geometry and action sequence visible before appearance synthesis. This separation helps distinguish changes in materials and lighting from changes in task structure. We inspect robot configuration, object arrangement, and viewpoint rather than expecting pixel-level agreement between the two representations. The later examples examine controllability by varying camera views or scene appearance while reusing the modeled task, and by applying targeted feedback and retrieved experience. Together, these examples complement the aggregate scores by showing which parts of the generation process can be inspected and revised.

\begingroup
\titlespacing*{\subsubsection}{0pt}{8pt}{4pt}
\setlength{\parskip}{4pt}

\noindent\begin{minipage}{\linewidth}
\textbf{Kitchen manipulation.} Figure\nobreakspace{}\ref{fig:example-01} pairs the arm and cabinet layout in the proxy model with a textured kitchen rendering. The confined workspace makes the relative placement of the end effector, drawer, and surrounding furniture important. The paired frames expose the geometry and action reference supplied to the appearance model.

\par\smallskip
\begingroup
\captionsetup{type=figure}
\centering
\includegraphics[width=.90\linewidth]{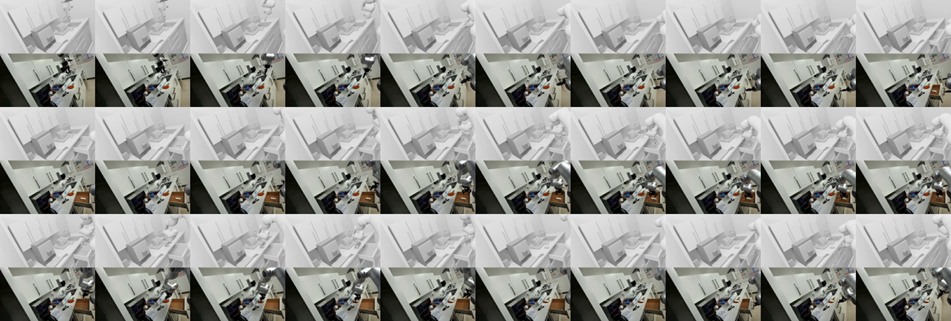}
\caption{Proxy and final rendered frames for kitchen drawer retrieval.}
\label{fig:example-01}
\par\endgroup

\end{minipage}\par\addvspace{10pt}

\noindent\begin{minipage}{\linewidth}
\textbf{Locomotion in a meeting room.} Figure\nobreakspace{}\ref{fig:example-02} shows a quadruped among chairs, tables, and walls. Furniture enters and leaves the frame as the view changes in both sequences. This example illustrates how a persistent scene and camera trajectory coordinate robot motion with changes in the visible background.

\par\smallskip
\begingroup
\captionsetup{type=figure}
\centering
\includegraphics[width=.90\linewidth]{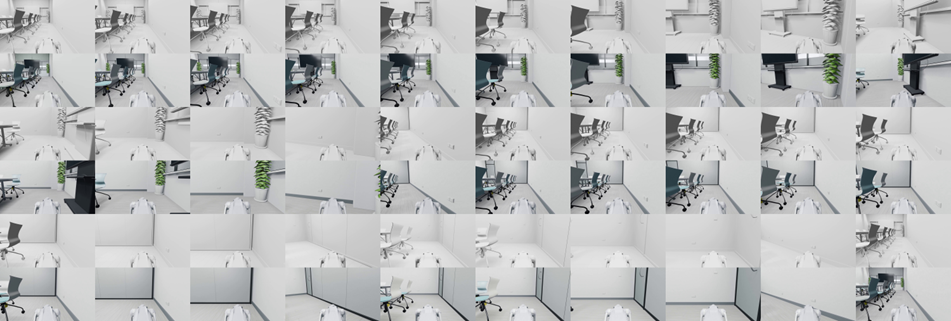}
\caption{Proxy and final rendered frames for quadruped locomotion in a meeting room.}
\label{fig:example-02}
\par\endgroup

\end{minipage}\par\addvspace{10pt}

\noindent\begin{minipage}{\linewidth}
\textbf{Quadruped locomotion in an office.} Figure\nobreakspace{}\ref{fig:example-03} extends the locomotion examples to a furnished office with windows and desks. The paired frames support inspection of robot visibility, passage through the scene, and viewpoint changes. The proxy model provides a spatial reference for interpreting the added materials and lighting.

\par\smallskip
\begingroup
\captionsetup{type=figure}
\centering
\includegraphics[width=.90\linewidth]{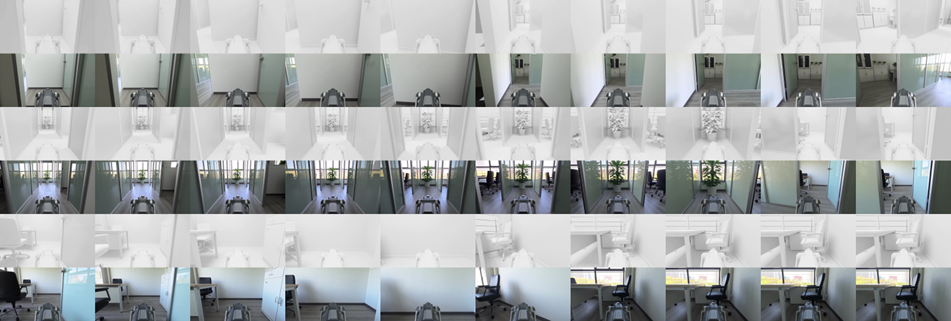}
\caption{Proxy and final rendered frames for quadruped locomotion in a furnished office.}
\label{fig:example-03}
\par\endgroup

\end{minipage}\par\addvspace{10pt}

\noindent\begin{minipage}{\linewidth}
\textbf{Humanoid locomotion in an office.} Figure\nobreakspace{}\ref{fig:example-04} shows a taller embodiment moving through an office. The robot occupies the image differently from the quadruped and interacts visually with furniture at a different height. This case highlights the role of robot scale, body pose, and camera framing in scene composition.

\par\smallskip
\begingroup
\captionsetup{type=figure}
\centering
\includegraphics[width=.90\linewidth]{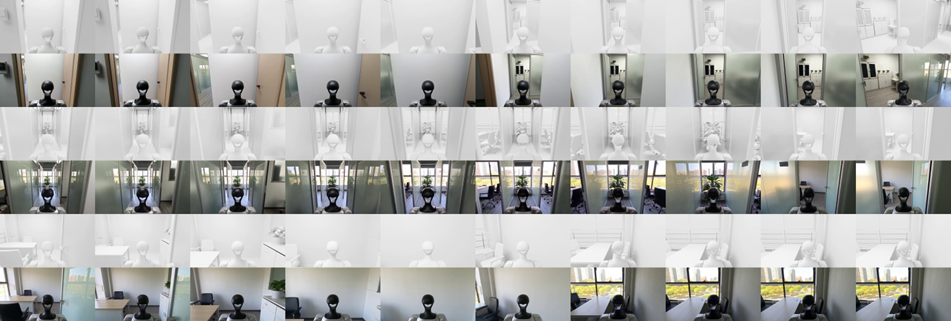}
\caption{Proxy and final rendered frames for humanoid locomotion in a furnished office.}
\label{fig:example-04}
\par\endgroup

\end{minipage}\par\addvspace{10pt}

\noindent\begin{minipage}{\linewidth}
\textbf{Coordinated dual-arm manipulation.} The filtration sequence in Figure\nobreakspace{}\ref{fig:example-05} emphasizes coordinated arm motion around a shared workspace. The proxy and final rendered rows expose arm configurations and the relative placement of the task objects over time. These relationships provide concrete visual checks of task structure as appearance is added.

\par\smallskip
\begingroup
\captionsetup{type=figure}
\centering
\includegraphics[width=.90\linewidth]{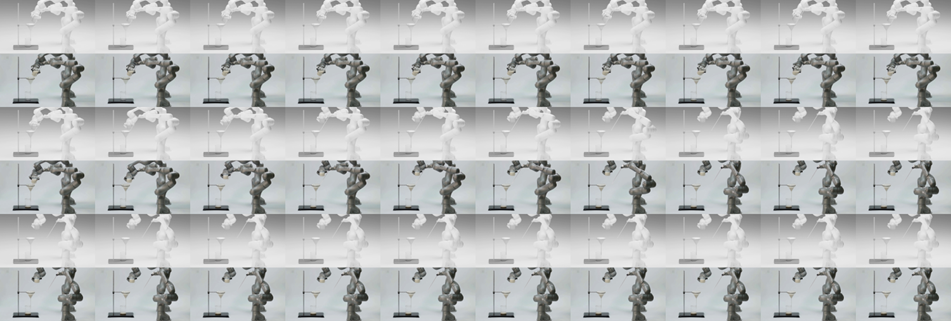}
\caption{Proxy and final rendered frames for dual-arm filtration.}
\label{fig:example-05}
\par\endgroup

\end{minipage}\par\addvspace{10pt}

\noindent\begin{minipage}{\linewidth}
\textbf{Dexterous manipulation.} Figure\nobreakspace{}\ref{fig:example-06} presents the suture task from a close viewpoint. The proxy frames expose hand configuration and workspace geometry, while the rendered rows add surface appearance. Fine finger motion and the thin suture remain difficult to assess from sampled frames alone, particularly for precise contact.

\par\smallskip
\begingroup
\captionsetup{type=figure}
\centering
\includegraphics[width=.90\linewidth]{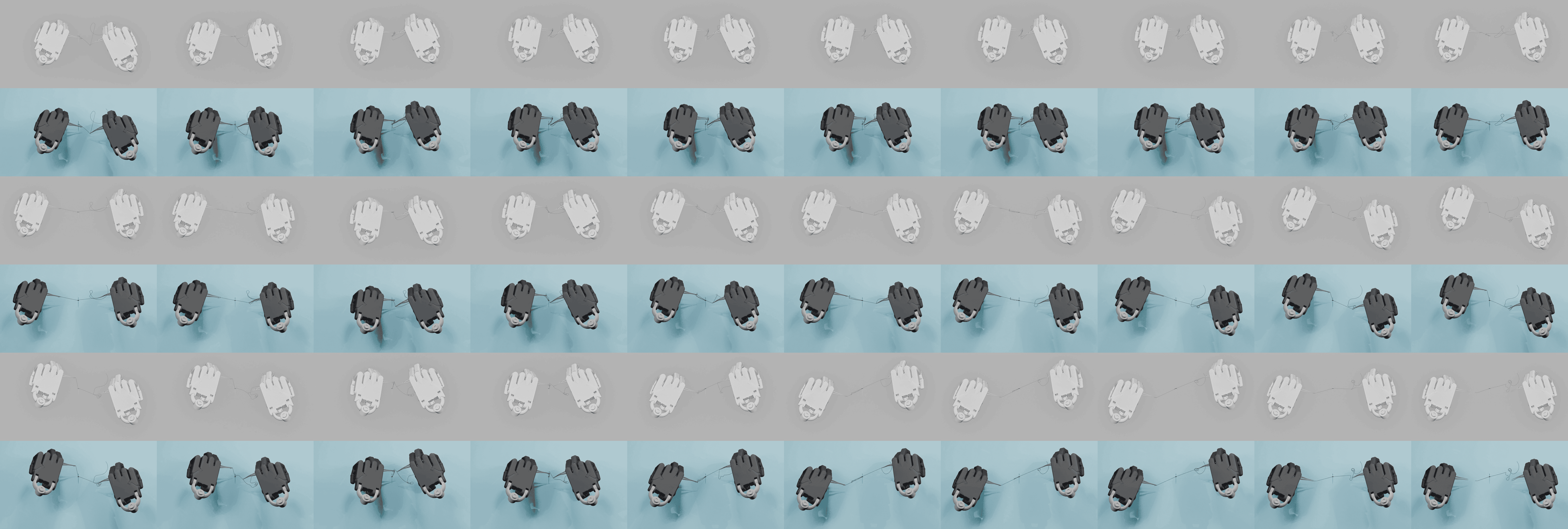}
\caption{Proxy and final rendered frames for dexterous-hand suture manipulation.}
\label{fig:example-06}
\par\endgroup

\end{minipage}\par\addvspace{10pt}

\noindent\begin{minipage}{\linewidth}
\textbf{A sequence of humanoid actions.} Figure\nobreakspace{}\ref{fig:example-07} follows a humanoid robot through door interaction and sitting down. The sequence combines locomotion, changes in body posture, and contact with scene objects. Comparing the paired rows helps trace these stages against a common scene layout while inspecting the appearance produced by the renderer.

The first row pair shows the approach to the door and the extension of an arm toward it. The middle pair follows the robot through the doorway and into the room, where the chair becomes a reference for the next action. In the final pair, the robot turns toward the chair and lowers its body into a seated posture. These transitions can be inspected through the robot's position relative to the door, chair, and nearby cabinets. The paired views also expose appearance changes around the same action stages, including added surface materials and plants in the rendered room.

Across these cases, the proxy videos provide an explicit reference for robot configuration, object placement, and camera motion. The final renderings add materials, lighting, and background detail. The montages illustrate the resulting visual correspondence; quantitative scores in Section\nobreakspace{}4.2 provide complementary evidence about consistency and instruction alignment.

\par\smallskip
\begingroup
\captionsetup{type=figure}
\centering
\includegraphics[width=.90\linewidth]{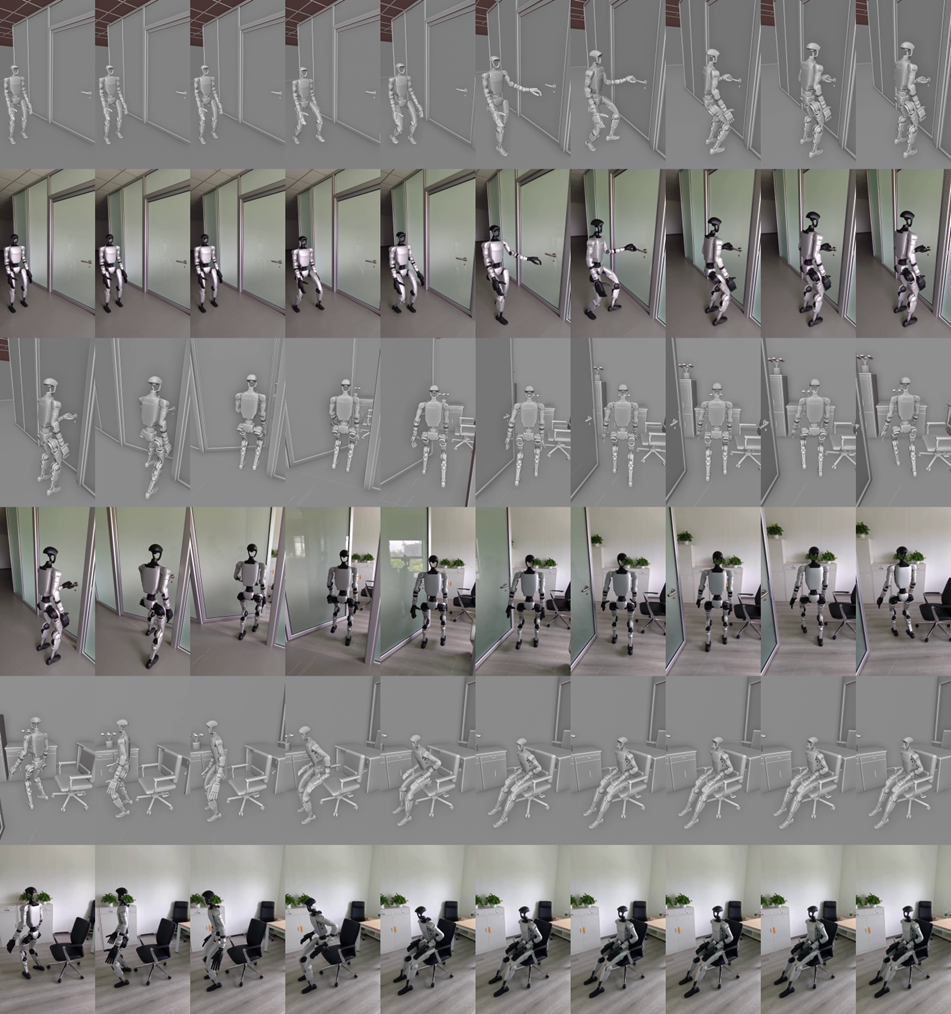}
\caption{Proxy and final rendered frames for a humanoid robot opening a door and sitting down.}
\label{fig:example-07}
\par\endgroup

\end{minipage}\par\addvspace{10pt}

\noindent\begin{minipage}{\linewidth}
\subsubsection{Multiple views of the same task}

Figure\nobreakspace{}\ref{fig:example-08} examines viewpoint control in a GO2W walking and stair-climbing task in a complex factory. Three cameras share the scene, robot trajectory, and timeline: the original view shows the robot relative to equipment and stairs; a following third-person view shows body motion and turning; and an approximate first-person view above the robot's front shows passage space and stair climbing. The added views share scene completion for newly visible regions. Each view yields a proxy video and a corresponding rendering.
\par\smallskip
\begingroup
\captionsetup{type=figure}
\centering
\includegraphics[width=.80\linewidth]{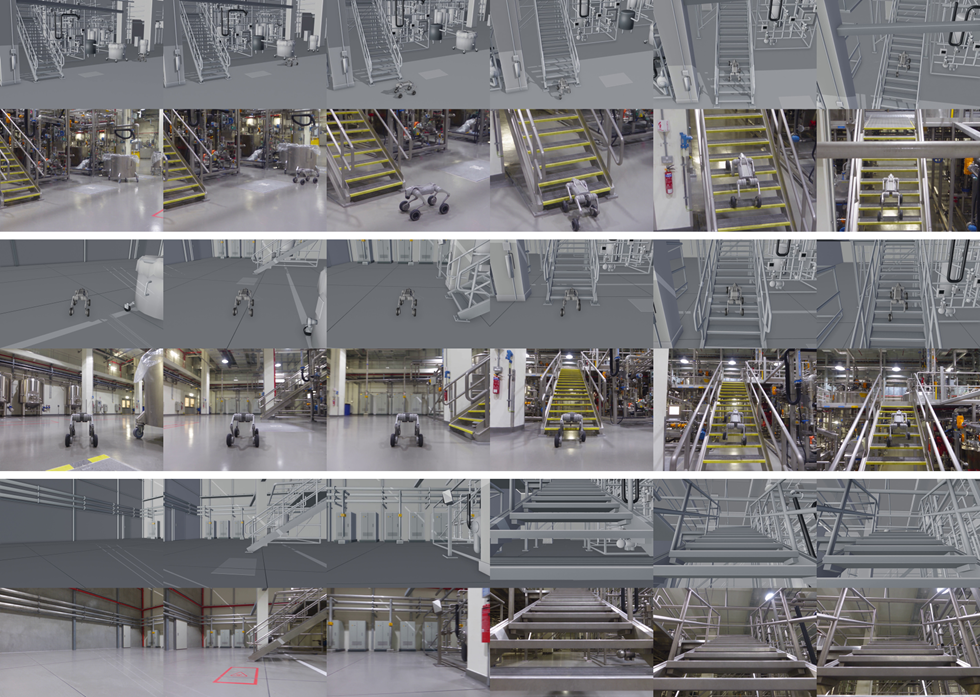}
\caption{Three views of the same GO2W factory task. The top, middle, and bottom pairs show the original, following third-person, and approximate first-person views, respectively.}
\label{fig:example-08}
\par\endgroup

\end{minipage}\par\addvspace{10pt}

\noindent\begin{minipage}{\linewidth}
\subsubsection{Multiple scene appearances from one proxy video}

In Figure\nobreakspace{}\ref{fig:example-09}, one proxy video of a humanoid transferring a box yields three appearances: a market with stalls and awnings, an industrial warehouse with a large hall and distant storage stacks, and a vertical farm with plant rows and supply equipment. All variants share the proxy video and reference frame. Prompts retain the robot, box, main shelves, camera, and action sequence while changing the environment during rendering. Shared geometry and motion thus support visual variation without rebuilding the Blender scene.
\par\smallskip
\begingroup
\captionsetup{type=figure}
\centering
\includegraphics[width=.80\linewidth]{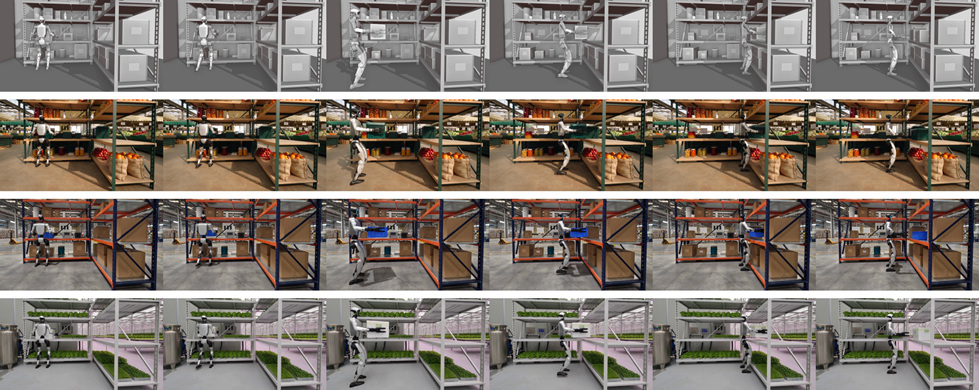}
\caption{Multiple scene appearances from one proxy video. From top to bottom: proxy model, market, industrial warehouse, and vertical farm.}
\label{fig:example-09}
\par\endgroup

\end{minipage}\par\addvspace{10pt}

\noindent\begin{minipage}{\linewidth}
\subsubsection{User feedback on modeling and rendering}

This laboratory case uses prewritten feedback to separate the mixing flask and receiving cup and replace one brief shake with three gentle swirls. The \mbox{BWM-Forge} Render pipeline then uses the fixed v2 proxy video; two further prompts address disappearing liquid and inconsistent color after pouring. Figure\nobreakspace{}\ref{fig:feedback} links the frames to these edits. Appendix\nobreakspace{}\ref{app:experience} provides full prompts and retrieved experience rules.
\par\smallskip
\begingroup
\captionsetup{type=figure}
\centering
\includegraphics[width=.80\linewidth]{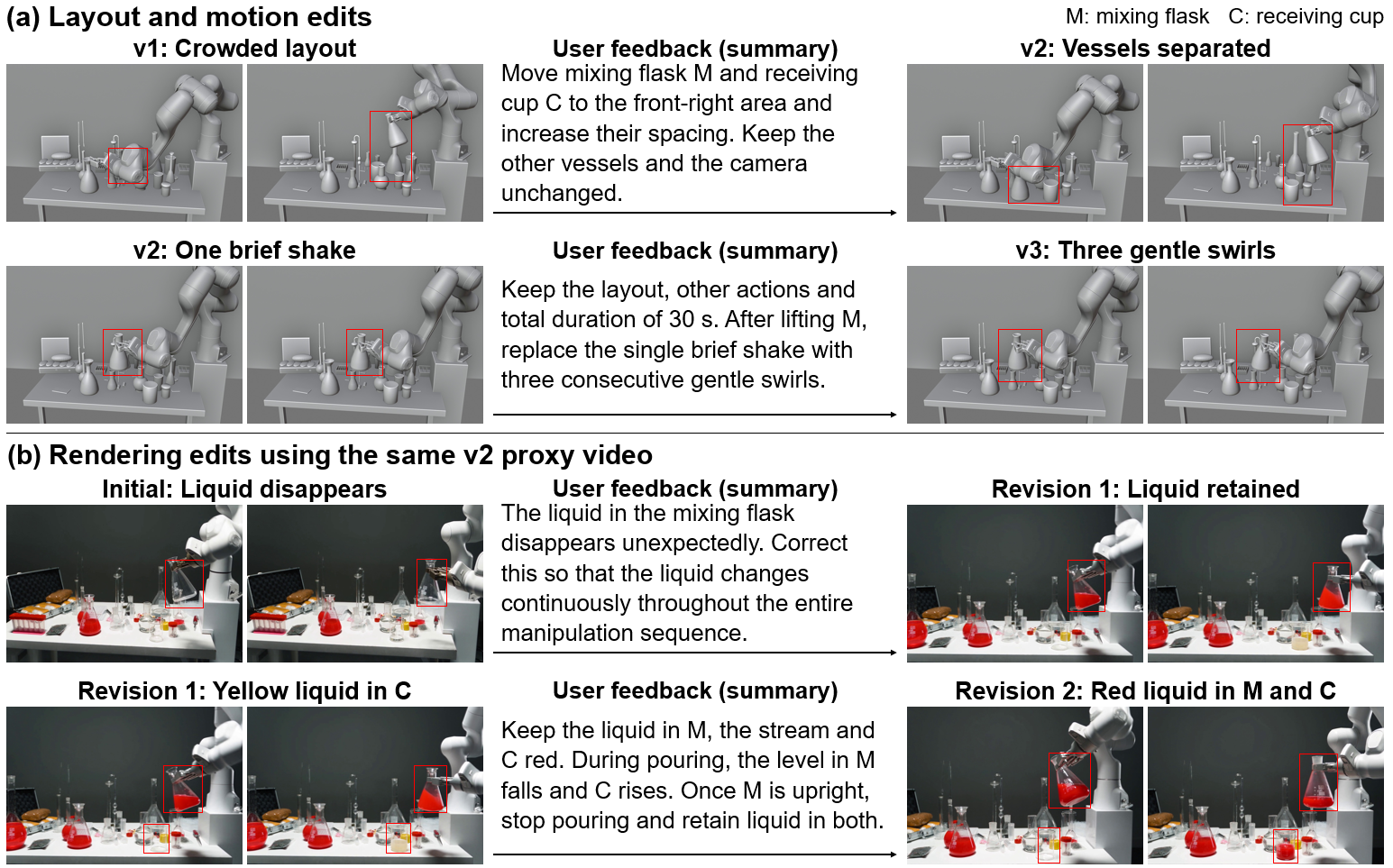}
\caption{User feedback on layout, motion, and appearance. Both rendering revisions share one proxy video.}
\label{fig:feedback}
\par\endgroup

\end{minipage}\par\addvspace{10pt}

\noindent\begin{minipage}{\linewidth}
\subsubsection{Harness experience reuse across scenes}

\mbox{BWM-Forge} builds, renders, and summarizes obstacle-transfer tasks in a workshop (A), laboratory (B), and kitchen (C). A starts without experience. B retrieves and applies four rules from A. C retrieves E1 to E3 from A and B, and E4 from A. Figure\nobreakspace{}\ref{fig:experience} shows rules for clearance, motion phases, container placement, and camera framing used with different object and obstacle dimensions.
\par\smallskip
\begingroup
\captionsetup{type=figure}
\centering
\includegraphics[width=.80\linewidth]{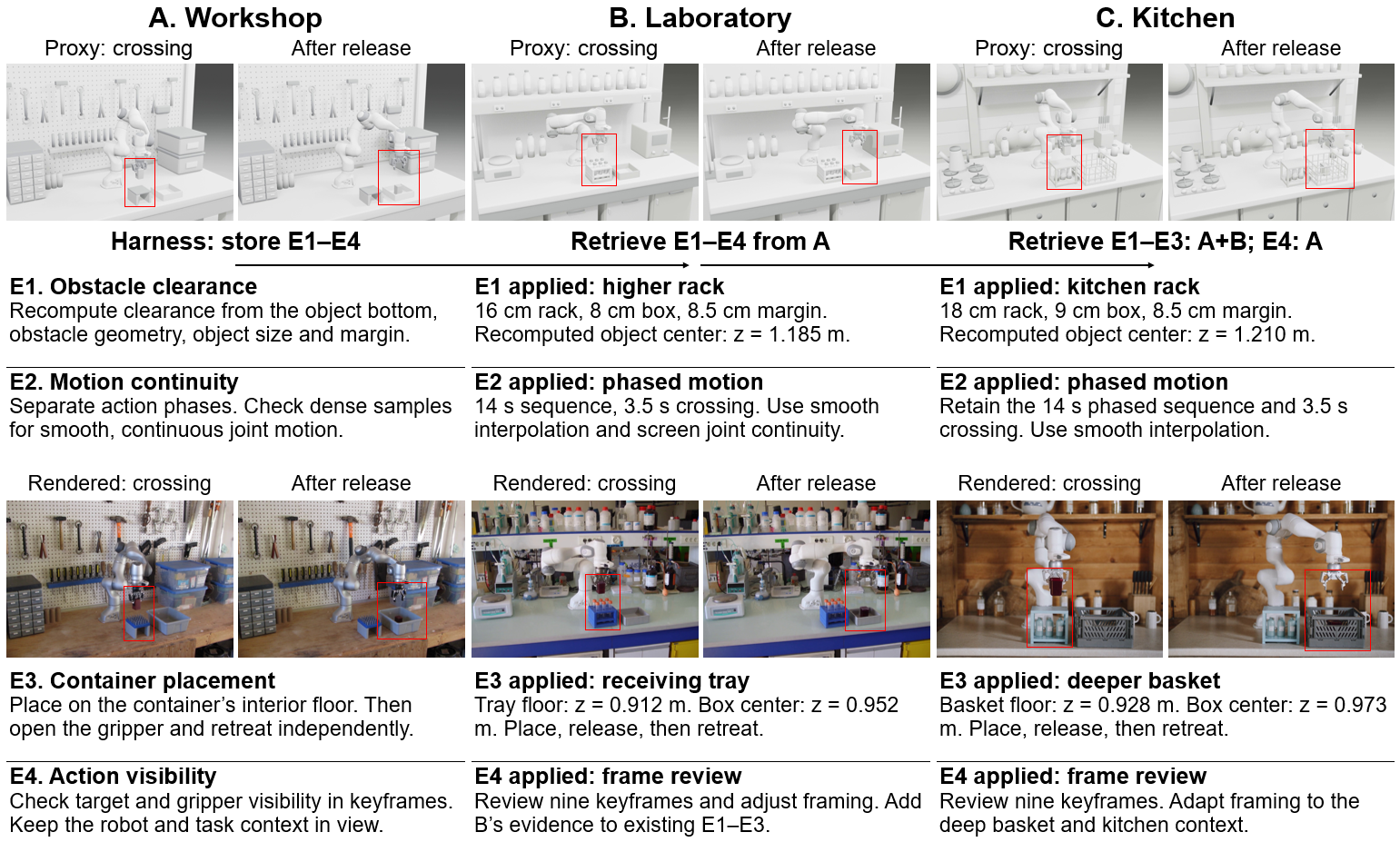}
\caption{Harness experience reuse from workshop A to laboratory B and kitchen C; full rules are in Appendix\nobreakspace{}\ref{app:experience}.}
\label{fig:experience}
\par\endgroup

\end{minipage}\par
\endgroup

\FloatBarrier
\section{Conclusion}
We presented \mbox{BWM-Forge}, a framework that uses executable scene programs to support controllable video generation for robotics. By separating scene geometry, robot motion, and camera control from appearance generation, \mbox{BWM-Forge} makes these variables accessible to inspection and revision. A multi-agent harness coordinates reconstruction, review, and rendering, while an experience repository supports reuse of modeling and rendering rules across tasks. The qualitative examples illustrate how this representation enables viewpoint changes, appearance variation, and targeted refinement through feedback.

On 15 tasks covering five robot types, \mbox{BWM-Forge} achieves higher best scores across the two evaluated video versions than \mbox{LingBot-World Base Cam} on all metrics except dynamic degree. These results support the potential of code-based representations for controllable robot video synthesis. The current evaluation is limited in scale and does not establish the utility of the generated videos for downstream robot learning. Broader task coverage, reduced reliance on manual feedback, and downstream policy evaluation are important next steps.

\section*{Contribution}
\noindent
\faUserTie\ 
\textcolor[HTML]{3F7FE8}{\textbf{Project Leader}}\par
\noindent Lanyun~Zhu\par

\medskip
\noindent
\faStar\ 
\textcolor[HTML]{E53935}{\textbf{Core Contributors}}\par
\noindent Chenming~Li, Jianhao~Song, Xuanyi~Liu, Zhixuan~Xu\par

\medskip
\noindent
\faUsers\ 
\textcolor[HTML]{29A34A}{\textbf{Contributors}}\par
\noindent
Yichen~An, Youyuan~Jiang, Haotian~Zhang, Ziming~Wei,
Enci~Xie, Hongye~Fang, Yueyang~Li, Zixuan~Zhang,
Xun~Jiang, Xuancheng~Xu, Zengrong~Lin, Baixu~Ji,
Chengxing~Xie, Yi~Guan, Zesi~Shen\par

\medskip
\noindent
\faEnvelope\ 
\textcolor[HTML]{7048B8}{\textbf{Corresponding Authors}}\par
\noindent
Lanyun~Zhu, Xing~Xu, Heng~Tao~Shen\par
\bibliographystyle{bwmforge}
\nocite{*}
\begingroup
\setlength{\bibsep}{1pt}
\bibliography{references}
\endgroup
\clearpage
\appendix
\section{VLM Visual Evaluation Prompt and Metrics}
\label{app:review}
\begingroup
\small
\setlength{\parskip}{4pt}
\subsection{VLM Scoring Prompt}

We used gpt-5.6-sol as our vision-language model (VLM), with the following prompt:

Generation task (data):

\$\{task\_prompt\}

Local objective scores (supporting statistics, not proof of task success):

\$\{objective\_scores\}

Appearance statistics (content-dependent supporting evidence):

\$\{appearance\_statistics\}

Review chronological proxy and generated samples independently, then compare. When supplied, the appearance image is a scoped appearance/initial-state reference; it cannot overwrite selected target structure or restore an excluded source entity.

Evaluate three metrics, each with an integer score from 0 to 100, visible evidence and a correction:

\begin{itemize}

\item visual\_realism: materials, lighting, shadows, geometry and generation artifacts; compare supported appearance where a reference is supplied.

\item physical\_temporal\_coherence: visible continuity, target structure and task-applicable support, contact and spatial relationships. Do not infer dynamic balance or real-device executability from images.

\item task\_semantic\_consistency: the task, selected target, required entities, event order, timing and state changes. Do not impose a particular robot anatomy or fixed behavioral sequence.

\end{itemize}

Complete all three action\_review checks. These are general checks, not mandatory action phases:

\begin{itemize}

\item initial: target identity, initial state and task-relevant arrangement.

\item task\_motion: required behavior, event order, target motion and applicable relationships across the sequence.

\item final\_state: required final state, outcome and retained entities. For ongoing or repetitive behavior, assess the required continuation at the end rather than inventing a terminal event.

\end{itemize}

Use pass/fail/uncertain with timestamps in both videos and specific visible evidence. Apply contact, grasp, support, release or travel checks only when required by the task and supported by actual modeled components. Do not require grasping or placement in an unrelated task. High technical scores or visual appeal cannot offset a failed task check. Mark occlusion, uninspected intervals or unsupported claims uncertain; do not invent a pass.

Comparison rows are paired by relative time, not detected events. Inspect surrounding samples before treating a pose difference as an event mismatch. Preserve source conflicts and upstream limitations.

Produce seedance\_correction\_prompt in at most 150 English words. Include only concrete, evidenced corrections; omit scores, metacommentary and unsupported style changes. Preserve the original task, unaffected content and scoped reference rules. Write all evidence, summaries and corrections in English and preserve the specified JSON field names and enums.

\subsection{VLM Evaluation Metrics}

\begin{center}
\begingroup
\small
\setlength{\tabcolsep}{4pt}
\renewcommand{\arraystretch}{1.15}
\begin{tabularx}{\linewidth}{@{}p{.23\linewidth}Xr@{}}
\toprule
Metric & Evaluation Criteria & Weight \\
\midrule
Visual realism & Whether materials, lighting, shadows, and the overall appearance resemble a real scene & 40\% \\
Physical and temporal coherence & Whether grasping, contact, motion, and changes between states are reasonable & 25\% \\
Task semantic consistency & Whether the video meets the task requirements and preserves key actions and object relationships & 35\% \\
\bottomrule
\end{tabularx}
\endgroup
\end{center}

\endgroup

\clearpage
\section{Full Prompts and Harness Experience}
\label{app:experience}
This appendix provides the complete user and author prompts for Figures\nobreakspace{}\ref{fig:feedback} and\nobreakspace{}\ref{fig:experience}, followed by the complete experience rules. Chinese prompts are translated into English. Repeated rules appear once, with their usage listed. The automatic common Render template, review-generated corrections, and log metadata are outside this collection of user prompts and rule bodies.

\subsection{Full prompts for user feedback}

\paragraph{P1  Initial task for Blender v1}

Use this image of a chemistry workbench. Keep the laboratory setting and the main vessels. Use a single robotic arm to perform a continuous task. First, pick up the conical flask containing red solution in the foreground. Pour part of the solution into a mixing flask that already contains liquid, then put the original flask back. Next, pick up the mixing flask and swirl it to mix the liquid. Pour the mixture into a receiving cup. Finally, put the flask down and withdraw the arm. Use a fixed camera to show the entire process clearly.

\paragraph{P2  Modeling feedback from v1 to v2}

The overall action sequence is present, but the mixing flask overlaps with the vessels and instruments in the center of the image. This makes the operating area difficult to see. Move the mixing flask and the final receiving cup to a less crowded area at the front right. Increase the spacing between them. Leave the other vessels where they are for now.

\paragraph{P3  Modeling feedback from v2 to v3}

The pouring and placement locations are now clear, but the mixing motion still consists of only one brief shake. Keep the rest of the sequence and the total duration unchanged. After picking up the mixing flask, gently swirl it three times in succession to make the mixing action more complete.

\paragraph{P4  First submitted Render feedback for v2}

In the current video, the liquid inside the second container disappears without a cause. This is the mixing flask that receives the first pour and is then picked up and swirled by the robotic arm. Please correct the continuity of the liquid throughout the operation.

\paragraph{P5  Second submitted Render feedback for v2}

Please further correct the continuity of liquid color and volume. M is the mixing flask that the robotic arm picks up second, briefly swirls, and then pours from. C is the final receiving cup. The liquid in M no longer disappears, but the red mixture turns pale yellow after it is poured into C. The changes in liquid volume before and after pouring are also unclear. Keep the mixture in M uniformly red. Keep the stream and the liquid received in C a similar shade of red. Do not make the liquid turn yellow or cloudy without a cause, or add a new chemical reaction effect. As M pours into C, the liquid level in M should gradually decrease and the liquid level in C should gradually increase. Stop the flow when M returns upright, and leave a small visible amount of liquid in M. The liquid in C should remain present during the subsequent placement and the final shot. Preserve the first pour from A into M, as well as the vessel positions, number, shapes, robotic arm motion, single brief swirl, action order, fixed camera, and timing specified by the current proxy video.

\subsection{Full prompts for experience reuse across scenes}

\paragraph{P6  Run A  complete task instruction}

On a maintenance workbench, use a Franka Panda with a Robotiq 2F-85 gripper to grasp the small cuboid part on the left, which is approximately 5 cm wide and 8 cm tall. Lift it vertically and carry it over the tool rack in the middle, which is approximately 10 cm tall. Place the part on the interior floor of the open storage box on the right. Open the gripper and retreat upward. Preserve the main structures in the reference image: the wooden workbench, rear tool pegboard, storage cabinet on the left, and stacked storage boxes on the right. The robot, part, tool rack, and receiving box are added for this task. Use a fixed third-person camera to show the complete approach, grasp, lift, obstacle crossing, placement, release, and retreat.

\paragraph{P7  Run A  scene supplement}

The reference image shows a maintenance work area viewed from the front left of the workbench. The main surface is a thick wooden tabletop. A white perforated pegboard at the rear holds various tools. Drawer storage and shelves are on the left. Two stacked storage boxes with blue lids are on the right. Place the added robotic arm and task objects in the free area at the front of the tabletop. Do not mirror the left and right structures. A single image does not provide absolute dimensions, occluded geometry, contact parameters, or dynamic parameters. These are explicitly specified for this model. Keyframe motion and geometric checks are not equivalent to validation through physical simulation.

\paragraph{P8  Run A  author-specified Render prompt}

Convert the given proxy video into a realistic and continuous video of robot manipulation. Use the reference image only to determine the scene style and background structures. The proxy video determines the camera, robot poses, object positions, actions, and their temporal order. Maintenance workshop: preserve the thick wooden tabletop, the light-colored perforated tool pegboard and tools at the rear, the storage drawers on the left, and the storage boxes with blue lids on the right. Use a dark red rigid material for the added part, a muted blue color for the middle tool rack, and gray plastic for the open receiving box. Give the background a realistic workshop appearance without adding moving objects.

Task: On a maintenance workbench, use a Franka Panda with a Robotiq 2F-85 gripper to grasp the small cuboid part on the left, which is approximately 5 cm wide and 8 cm tall. Lift it vertically and carry it over the tool rack in the middle, which is approximately 10 cm tall. Place the part on the interior floor of the open storage box on the right. Open the gripper and retreat upward. Preserve the main structures in the reference image: the wooden workbench, rear tool pegboard, storage cabinet on the left, and stacked storage boxes on the right. The robot, part, tool rack, and receiving box are added for this task. Use a fixed third-person camera to show the complete approach, grasp, lift, obstacle crossing, placement, release, and retreat.

Keep the structure of the Franka Panda arm and Robotiq two-finger gripper stable. Keep the target object's shape, dimensions, and identity unchanged throughout. After the gripper closes, keep the target in contact with the gripper. Do not pass through the obstacle during lifting and crossing. During placement, bring the object into contact with the container's interior floor. After the gripper opens, keep the target inside the container while the gripper retreats independently. Do not omit, repeat, or reverse actions. Do not add objects or abruptly change the target's position. Preserve the proxy video duration and fixed third-person composition. Keep the background stationary. Do not add subtitles, text labels, watermarks, split screens, camera cuts, or slow motion.

\paragraph{P9  Run B  complete task instruction}

On a laboratory workbench, use the same Franka Panda with a Robotiq 2F-85 gripper to grasp the closed cuboid sample box on the left. Lift it and carry it over the taller test-tube rack in the middle. Place it in the open receiving tray on the right. Open the gripper and retreat. Keep the sample box closed throughout the task. Do not perform any liquid manipulation. Recalculate the grasp and obstacle-crossing heights from the current object dimensions and rack height. Do not directly copy the absolute coordinates from the previous scene. Preserve the laboratory bench, rear reagent-bottle shelves, and cabinets below the bench in the reference image. Use a fixed third-person camera to show the complete action.

\paragraph{P10  Run B  scene supplement}

The reference image shows a laboratory workbench from the front and slightly above. Elevated shelves at the rear hold several capped bottles. Instruments are on the workbench, with cabinet doors and yellow panels below. The robot, closed sample box, test-tube rack used as an obstacle, and receiving tray are added for this task. Keep the work area visually separated from the bottle shelves in the background, and leave room for the gripper. Object dimensions, materials, and the robotic arm mounting position are modeling assumptions. The sample contents are unknown. Do not infer chemical substances from the photograph.

\paragraph{P11  Run B  author-specified Render prompt}

Convert the given proxy video into a realistic and continuous video of robot manipulation. Use the reference image only to determine the scene style and background structures. The proxy video determines the camera, robot poses, object positions, actions, and their temporal order. A tidy research laboratory with rich detail: a light gray chemically resistant worktop, multiple levels of reagent-bottle shelves at the rear, capped bottles, instruments, and cabinets below the bench. The target sample box is a closed dark red plastic box, the obstacle test-tube rack is blue, and the receiving tray is gray. Do not add lid opening, pouring, liquid flow, or extra robotic arm actions.

Task: On a laboratory workbench, use the same Franka Panda with a Robotiq 2F-85 gripper to grasp the closed cuboid sample box on the left. Lift it and carry it over the taller test-tube rack in the middle. Place it in the open receiving tray on the right. Open the gripper and retreat. Keep the sample box closed throughout the task. Do not perform any liquid manipulation. Recalculate the grasp and obstacle-crossing heights from the current object dimensions and rack height. Do not directly copy the absolute coordinates from the previous scene. Preserve the laboratory bench, rear reagent-bottle shelves, and cabinets below the bench in the reference image. Use a fixed third-person camera to show the complete action.

Keep the structure of the Franka Panda arm and Robotiq two-finger gripper stable. Keep the target object's shape, dimensions, and identity unchanged throughout. After the gripper closes, keep the target in contact with the gripper. Do not pass through the obstacle during lifting and crossing. During placement, bring the object into contact with the container's interior floor. After the gripper opens, keep the target inside the container while the gripper retreats independently. Do not omit, repeat, or reverse actions. Do not add objects or abruptly change the target's position. Preserve the proxy video duration and fixed third-person composition. Keep the background stationary. Do not add subtitles, text labels, watermarks, split screens, camera cuts, or slow motion.

\paragraph{P12  Run C  complete task instruction}

On a kitchen counter, use the same Franka Panda with a Robotiq 2F-85 gripper to grasp the lidded cuboid food-storage box on the left. Lift it and carry it over the kitchen storage rack in the middle. Place it in the deeper open storage basket on the right. Plan the clearance from the dimensions of the current box, obstacle, and basket opening. Lower the box onto the interior floor of the basket. Open the gripper and retreat along a collision-free direction. Keep the lid closed throughout the task. Preserve the wooden wall, rear shelves, and kitchen items in the reference image. Use a fixed third-person camera to show the complete action.

\paragraph{P13  Run C  scene supplement}

The reference image shows a kitchen with warm wood finishes. The wall and cabinets are mainly wood-colored. Shelves are above, a stove and kettle are on the left, and seasonings and kitchen utensils are at the rear. The robotic arm, lidded food box, obstacle rack, and receiving basket are added for this task and placed in the front work area. Reconstruct only the broad visible layout. Absolute dimensions and occluded structures are specified during modeling. The height of the basket's interior floor differs from that of the surrounding counter, and the two must be recorded separately.

\paragraph{P14  Run C  author-specified Render prompt}

Convert the given proxy video into a realistic and continuous video of robot manipulation. Use the reference image only to determine the scene style and background structures. The proxy video determines the camera, robot poses, object positions, actions, and their temporal order. A warm modern kitchen: wooden wall panels, open shelves at the rear, a stationary kettle, seasoning bottles, and tableware. The countertop is clean, light-colored stone. The moving target is a dark red lidded food-storage box. The obstacle storage rack is muted blue, and the deeper receiving basket is gray. Do not open the lid, turn the target into food, or add cooking actions.

Task: On a kitchen counter, use the same Franka Panda with a Robotiq 2F-85 gripper to grasp the lidded cuboid food-storage box on the left. Lift it and carry it over the kitchen storage rack in the middle. Place it in the deeper open storage basket on the right. Plan the clearance from the dimensions of the current box, obstacle, and basket opening. Lower the box onto the interior floor of the basket. Open the gripper and retreat along a collision-free direction. Keep the lid closed throughout the task. Preserve the wooden wall, rear shelves, and kitchen items in the reference image. Use a fixed third-person camera to show the complete action.

Keep the structure of the Franka Panda arm and Robotiq two-finger gripper stable. Keep the target object's shape, dimensions, and identity unchanged throughout. After the gripper closes, keep the target in contact with the gripper. Do not pass through the obstacle during lifting and crossing. During placement, bring the object into contact with the container's interior floor. After the gripper opens, keep the target inside the container while the gripper retreats independently. Do not omit, repeat, or reverse actions. Do not add objects or abruptly change the target's position. Preserve the proxy video duration and fixed third-person composition. Keep the background stationary. Do not add subtitles, text labels, watermarks, split screens, camera cuts, or slow motion.

\subsection{Full experience rules retrieved in the feedback case}

The following five rules came from the existing experience store. All five were included in the initial Render prompts for Blender v1, v2, and v3. The first v2 rendering revision retained four rules, and the second retained three.

\paragraph{H1  Upstream provenance and separate modifications}

Rule ID: 4d929eaf57fe0ea157de. Used in all three initial renderings and both v2 rendering revisions.

Build the composition from a read-only upstream project. Preserve source hashes, role markers, and unresolved assumptions. Organize first-frame registration, motion retargeting, and contact geometry as separate modifications.

\paragraph{H2  Geometry inspection and evidence limits}

Rule ID: 994c65b3f9070a1f043c. Used in all three initial renderings and both v2 rendering revisions.

Combine inspection of the full triangle meshes, contact rays, support rays, and checks of key poses. When local sampling or an AABB sweep finds no issues, report only that no issues were found within the sampled region. Do not claim continuous collision freedom or physical success.

\paragraph{H3  Joint and temporal continuity}

Rule ID: 3987ed7ebab6d60c9451. Used in all three initial renderings and both v2 rendering revisions.

Emphasizing smooth motion in the prompt alone is insufficient to resolve abrupt motion changes. Use more explicit joint-level continuity, contact constraints, and temporal consistency controls together.

\paragraph{H4  Motion-channel preservation and retargeting}

Rule ID: 5e12b5ffc277b5d2352b. Used in all three initial renderings.

Preserve the original motion channels and sampling semantics without modification, and separately record the retargeted joint curves used to complete the task. Do not directly interpret uncalibrated gripper values as actual closing distances, and do not silently clip original values that exceed limits.

\paragraph{H5  Hollow-container geometry}

Rule ID: 3cb6158e6115f84d2904. Used in all three initial renderings and the first v2 rendering revision.

Model containers with genuine hollow interiors. Retain a bottom panel, side panels, a rear panel, and adjustable wall thickness. Use opening rays or similar geometric checks to confirm that the target region actually falls on the interior floor. Do not substitute a closed solid body.

\subsection{Full experience rules retrieved across scenes}

\paragraph{E1  object-based obstacle clearance}

Rule ID: 705c934f96e00a57bbac.

When carrying a rigid object with a fixed pose over an obstacle whose position and height are known, calculate clearance from the object's lowest point rather than from the end-effector center. Plan distinct vertical lift, constant-height crossing, and descent phases; recompute the crossing height from the current obstacle top, object extent, and required margin, and inspect object clearance throughout the obstacle region. Do not reuse an absolute height from another scene, and do not treat sampled geometric or kinematic clearance as continuous collision, contact, dynamics, or real-robot validation.

\paragraph{E2  motion continuity and visible phases}

Rule ID: fdc32c82e1e6a5b947af. Stored scope: robot = franka\_panda\_robotiq\_2f85; task\_type = obstacle\_transfer\_place.

For the documented Franka Panda and Robotiq 2F-85 obstacle-transfer placement workflow, prioritize a smooth trajectory and make grasp closure, vertical lift, obstacle crossing, descent to the container floor, release, and upward retreat visually distinct. Verify continuity with sufficiently dense temporal evidence; do not infer uninterrupted grasp, collision-free clearance, exact release, or physical validity from sparse sampled poses, a correct final deposited state, process completion, or later archival confirmation. A retained render with a RETRY decision remains quality-limited unless the flagged motion issue is separately resolved.

\paragraph{E3  interior support and separate release}

Rule ID: faa8bd0672a30109e04a.

When placing a rigid object into an open container with a defined interior floor, model and record the interior support-surface height separately and place the object's bottom on that surface rather than assuming the surrounding tabletop height. Separate placement, gripper opening, and gripper retreat into distinct intervals so the final object can remain in the container while the gripper leaves independently. A matched modeled support height and visually correct final frame do not by themselves validate contact dynamics or physical task success.

\paragraph{E4  action visibility}

Rule ID: 8ad3f56f09610a6e13d7.

For fixed third-person videos of tabletop small-object manipulation, render representative grasp, transfer or obstacle-crossing, and placement frames before exporting the full video, and check the on-screen size and visibility of both the target and gripper. If irrelevant floor or table-leg regions dominate or the target is hard to distinguish, tighten the framing around the work area while retaining the complete robot and task-relevant scene structures. Treat this as an action-visibility check, not evidence of physical task success.

\paragraph{Additional Render rule  Gripper geometry and object attachment}

Rule ID: 47de34d1b9f94acd0c99. Stored scope: robot = franka\_panda\_robotiq\_2f85.

For the verified Franka Panda with Robotiq 2F-85 asset when parallel-finger grasping a size-compatible regular rigid object, derive closed-finger spacing and TCP offset from the actual inner-pad geometry. Bind object motion to the verified end-effector transform only after the modeled closing phase has completed; after the object reaches its modeled support surface, end the binding, open the fingers, and retreat independently. Recalculate object width, finger fit, TCP offset, and source and destination support heights for every new object. These animation constraints require the cited asset and do not establish contact dynamics, real-robot precision, or physical task success.

\clearpage
\section{Evaluation Configuration}
\renewcommand{\thetable}{C.\arabic{table}}
\setcounter{table}{0}
\subsection{Video Resolutions}

Table\nobreakspace{}\ref{tab:resolutions} lists the resolutions of the evaluation videos. LingBot-World output resolution depends on the aspect ratio of its reference image.

\begin{table}[htbp]
\centering
\caption{Resolutions of the three groups of evaluation videos.}
\label{tab:resolutions}
\begingroup
\small
\setlength{\tabcolsep}{4pt}
\renewcommand{\arraystretch}{1.15}
\begin{tabularx}{\linewidth}{@{}Xrrr@{}}
\toprule
\textbf{Task} & \textbf{\mbox{BWM-Forge}} & \shortstack{\textbf{\mbox{BWM-Forge}}\\\textbf{16 FPS}} & \textbf{LingBot-World} \\
\midrule
Kitchen drawer retrieval & 1280\ensuremath{\times}720 & 1280\ensuremath{\times}720 & 832\ensuremath{\times}464 \\
Meeting room quadruped & 1280\ensuremath{\times}720 & 1280\ensuremath{\times}720 & 720\ensuremath{\times}544 \\
Office quadruped & 1920\ensuremath{\times}1080 & 1920\ensuremath{\times}1080 & 464\ensuremath{\times}832 \\
Office humanoid robot & 1280\ensuremath{\times}720 & 1280\ensuremath{\times}720 & 464\ensuremath{\times}832 \\
Dual arm filtration & 1280\ensuremath{\times}720 & 1280\ensuremath{\times}720 & 832\ensuremath{\times}464 \\
Dexterous hand suture manipulation & 1280\ensuremath{\times}720 & 1280\ensuremath{\times}720 & 832\ensuremath{\times}464 \\
Humanoid robot opens a door and sits down & 720\ensuremath{\times}1280 & 720\ensuremath{\times}1280 & 464\ensuremath{\times}832 \\
\bottomrule
\end{tabularx}
\endgroup
\end{table}

\subsection{Frame Rate Conversion}
\label{app:frame-rate}

The 16 FPS versions preserve the spatial resolution and playback speed. They are created by frame sampling without motion interpolation and encoded with libx264, preset=medium, CRF=18, and yuv420p. This conversion changes both frame rate and encoding.

\subsection{LingBot-World Configuration}

All seven LingBot-World videos use seed 42, 70 sampling steps, 161 frames, and 16 FPS. Generation uses the i2v-A14B entry point with size=480\ensuremath{\times}832, the UniPC sampler, sample\_shift=10.0, and guide\_scale=(5.0, 5.0). The model runs on eight NVIDIA A100 40 GB GPUs with FSDP and Ulysses=8.

The environment uses Python 3.10.20, PyTorch 2.7.1, CUDA 12.6, FlashAttention 2.8.3.post1, Transformers 4.51.3, and Diffusers 0.39.0. Outputs are H.264 MP4 videos with the yuv420p pixel format. Each video lasts 10.0625 s.

\end{document}